\documentclass[10pt,twocolumn,letterpaper]{article}

\usepackage[pagenumbers]{wacv} 

\definecolor{wacvblue}{rgb}{0.21,0.49,0.74}
\usepackage[pagebackref,breaklinks,colorlinks,allcolors=wacvblue]{hyperref}
\usepackage{multirow}
\usepackage{tabularx}
\usepackage[table]{xcolor}
\usepackage{caption}
\usepackage{arydshln}
\usepackage{float}
\definecolor{cSourceOnly}{RGB}{217,94,79}   
\definecolor{cAvg}{RGB}{115,158,207}        
\definecolor{cGlobal}{RGB}{84,168,105}      
\definecolor{cTargetOnly}{RGB}{242,194,77}  
\usepackage[accsupp]{axessibility}  

\def\wacvPaperID{79} 
\def\confName{WACV}
\def\confYear{2027}

\title{How Far from Clinical Deployment? Evaluating the Complete Unsupervised Domain Adaptation Pipeline in Medical Imaging}

\author{
Yiheng Xiong$^{1}$ \quad Luisa Gall\'ee$^{1}$ \quad Daniel Santak Wolf$^{1,2}$ \quad Heiko Hillenhagen$^{1}$ \quad Michael G\"otz$^{1}$\\
\small$^{1}$Section of Experimental Radiology, Ulm University Medical Center \\ \small$^{2}$Visual Computing Group, Ulm University
}

\begin{document}

\maketitle
\begin{abstract}
Deploying unsupervised domain adaptation (UDA) in clinical practice requires choosing which algorithm to use and which of its trained models to ship. However, the deployment (target) domain is unlabeled, so models cannot be evaluated directly on it, leaving it unclear which to select. We address this by evaluating the complete UDA pipeline, considering both adaptation and label-free selection together. Our study covers eleven clinically relevant cross-domain scenarios from nine medical imaging datasets, with ten UDA algorithms and 13 label-free selection methods (validators), evaluating over 80{,}000 trained models in total. By this, we find that a capable adapted model usually exists, but identifying it without target labels is difficult: the validator-selected models leave a large and structural target performance gap to the best available one, with no evaluated validator consistently reliable. Towards closing it, we explore two strategies, ensembling and a small target-labeling budget; both narrow this gap but do not close it entirely. Overall, deployable UDA depends on the complete pipeline; addressing the less explored selection step could bring much of current UDA closer to clinical use. Our code is open sourced at \href{https://xiongyiheng.github.io/complete-uda-pipeline-site/}{Project Page}.
\end{abstract}

\section{Introduction}
\label{sec:intro}
\begin{figure}[h]
    \centering
    \includegraphics[width=1.0\linewidth]{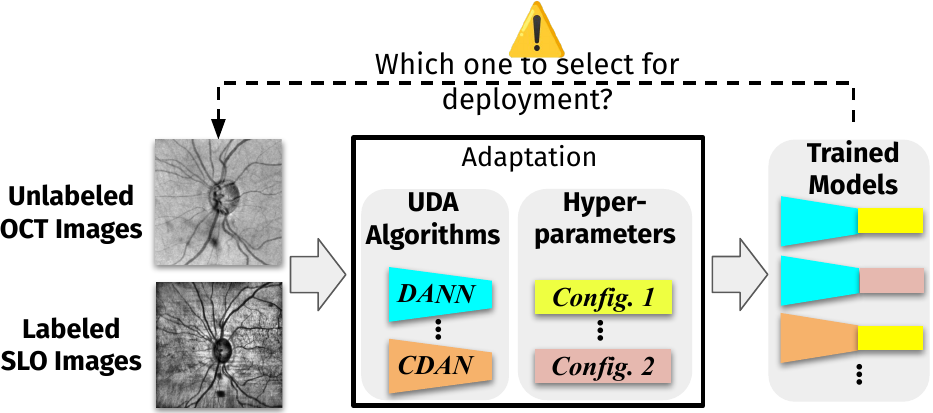}
    \caption{The complete UDA pipeline at deployment. Adaptation trains models on a labeled source domain (here SLO) and an unlabeled target domain (here OCT), producing many candidate models that vary in UDA algorithm and hyperparameter configuration. At deployment, one model needs to be selected, yet without target labels they cannot be evaluated directly to decide which.}
\label{fig:teaser}
\end{figure}

Deep learning models excel at medical image-based diagnosis, but their performance often degrades substantially under domain shifts arising from differences in imaging modalities, acquisition protocols, and patient populations~\cite{kumari2024deep}.
Annotating training data from a new domain is costly, requiring scarce medical expertise. Unsupervised domain adaptation (UDA) addresses both challenges by transferring knowledge from a labeled source domain to an unlabeled target domain, with recent algorithms adapting robustly across clinically relevant shifts, such as cross-modality~\cite{xie2022unsupervised}, cross-site~\cite{liu2025hasd}, and
cross-age~\cite{feng2023contrastive}. However, deploying UDA in clinical routine requires a practitioner to decide which algorithm to use and which of its checkpoints (a model snapshot at a training iteration) to ship, as shown in Figure~\ref{fig:teaser}. Thus, model selection is an unavoidable part of the complete UDA pipeline. Furthermore, without target labels, candidate models cannot be evaluated directly on the target domain, leaving it unclear which one to select for deployment: the standard approach of comparing them on a held-out labeled validation set is unavailable~\cite{musgrave2021unsupervised}. Yet, existing medical UDA surveys~\cite{choudhary2020advancing,guan2021domain,sarafraz2022domain,li2023medical,kumari2024deep,su2024navigating,yang2026survey} and evaluations~\cite{dorent2023crossmoda,pu2024m3,chamarthi2024mitigating,shirokikh2025m3da,sultana2025domain} primarily focus on adaptation in isolation, leaving the selection step largely unexamined. As a result, it remains unclear how the complete UDA pipeline behaves under clinical deployment conditions, and thus how far current UDA truly is from clinical deployment.

To address this, we evaluate the complete UDA pipeline in medical imaging, considering both adaptation and label-free selection that clinical deployment requires. For adaptation, we include ten UDA algorithms spanning multiple paradigms, with some tailored to medical imaging~\cite{guan2021multi,zhang2020collaborative}. For selection, we adopt 13 established label-free selection approaches (a.k.a. validators), such as IWCV~\cite{sugiyama2007covariate} and DEV~\cite{you2019towards}. Each validator assigns a scalar validation score to each checkpoint without using target labels, and the checkpoint with the best score is selected. Our evaluation covers brain MRI, chest X-ray (CXR), and retinal imaging, with eleven clinically relevant cross-domain scenarios in total. Brain MRI and CXR use four datasets each, with within-modality shifts arising from differences in hospitals, scanners, or patient populations, while retinal imaging involves across-modality shifts between SLO and OCT. Altogether, this amounts to roughly 16{,}500 checkpoint configurations (each set by
the algorithm, its hyperparameters, and training iteration), or over 80{,}000 trained checkpoints once repeated across folds or random seeds.

Our study points to a consistent picture across different cross-domain scenarios. Adaptation itself is often effective: a capable adapted model usually exists among the candidates. What proves difficult is identifying it without target labels: the models selected by the evaluated validators leave a large target performance gap to the best available model. We further trace the origin of this gap and find it largely structural: no single evaluated validator works consistently well across all scenarios; one that succeeds in some fails in others, so no default choice can be recommended in advance. In the absence of a consistently reliable validator across all scenarios, towards closing the gap, we turn to two strategies. Instead of relying on one validator, the first ensembles predictions across validator-selected checkpoints from every algorithm; instead of remaining fully label-free, the second spends a small target-labeling budget. Both narrow the gap, though neither closes it entirely and each carries its own limitations.

Our contributions are summarized as follows:
\begin{itemize}
\item We present, to our knowledge, the first study to evaluate the complete adaptation and label-free selection UDA pipeline in medical imaging under clinical deployment conditions, spanning ten algorithms, 13 validators, and nine datasets across eleven cross-domain scenarios.

\item By considering adaptation and selection as a whole, we find that a capable adapted model usually exists, yet the remaining distance to clinical deployment lies largely in label-free selection: the selection gap is large and structural, with no evaluated validator consistently reliable.

\item Towards closing the gap, we point to two strategies, ensembling and a small target-labeling budget; both narrow this gap and open directions for more reliable selection.
\end{itemize}

\section{Related Work}
\label{sec:related_work}
\subsection{Surveys and Benchmarks on Medical UDA}
UDA in medical imaging has been reviewed from several angles. Choudhary et al.~\cite{choudhary2020advancing} give an early overview of deep domain adaptation, Guan and Liu~\cite{guan2021domain} provide a widely used taxonomy across modalities,
and Kumari and Singh~\cite{kumari2024deep} survey more recent deep UDA methods and
datasets. More recent surveys cover emerging diffusion- and foundation-model
approaches~\cite{yang2026survey}, broader reviews of distribution shift also include
UDA~\cite{su2024navigating}, and others focus on specific settings such as functional
brain data~\cite{sarafraz2022domain} and segmentation~\cite{li2023medical}. Alongside
these surveys, several benchmarks evaluate adaptation under realistic shifts:
M3DA~\cite{shirokikh2025m3da} for 3D MRI/CT segmentation,
CrossMoDA~\cite{dorent2023crossmoda} for cross-modality segmentation,
M3-UDA~\cite{pu2024m3} for multi-hospital fetal ultrasound detection, and Chamarthi
et al.~\cite{chamarthi2024mitigating} and Sultana et al.~\cite{sultana2025domain} for
skin lesion classification. These efforts have established rigorous comparisons of
adaptation methods on clinically relevant data, but the selection step remains
largely underexplored. Our study complements them by evaluating the complete pipeline
that clinical deployment requires, adaptation and label-free selection together.

\subsection{Label-Free Model Selection in UDA}
Selecting a model without target labels has been studied mainly in general-vision UDA, where a range of validators have been
proposed~\cite{sugiyama2007covariate,ganin2015unsupervised,you2019towards,saito2021tune,musgrave2022three,tu2024does,yang2024can,hu2023mixed} and evaluated. Musgrave et al.~\cite{musgrave2021unsupervised} provide one of the first UDA model selection benchmarks covering many algorithms, though over a relatively small set of three label-free validators. Subsequent evaluations~\cite{ericsson2023better,hu2024towards,lalou2024skada} strengthen evaluation practice further with more validators, with SKADA-Bench~\cite{lalou2024skada} additionally covering broader modalities, though with an emphasis on shallow adaptation methods. These works, however, primarily focus on selecting checkpoints within individual algorithms. In practice, deployment also requires choosing which algorithm to use, so
a more realistic setting also calls for selecting across algorithms. Beyond this, medical images pose distinct challenges, exhibiting more pronounced and varied shifts than natural images~\cite{kumari2024deep}, under which the behavior of these validators remains unclear. To address these, our study examines selection also across algorithms, over a broad set of 13 validators and ten deep UDA algorithms, including medical-specific ones, and, in
particular, under the shifts that clinical routine faces, such as cross-modality, cross-institution, and cross-cohort, assessing how the complete UDA pipeline behaves under deployment conditions.

\section{Datasets and Methods}
\label{sec:dataset_and_method}
\begin{table*}[htbp]
\centering
\caption{Overview of the datasets, UDA algorithms, and validators used in our study. \textbf{(a)} Datasets across brain MRI, CXR, and retinal imaging (SLO / OCT), with the number of positive (Pos.) and negative (Neg.) samples per dataset. Positive and negative denote, respectively, Alzheimer's disease and cognitively normal for brain MRI, pneumonia and non-pneumonia for CXR, and glaucoma and non-glaucoma for retinal imaging. \textbf{(b)} UDA algorithms spanning multiple paradigms. \textbf{(c)} Validators grouped into source-guided and target-based criteria.}
\label{tab:benchmark_overview}
\footnotesize
\begin{minipage}[t]{0.40\linewidth}
\centering
\textbf{(a) Datasets}\\[0.01cm]
\setlength{\tabcolsep}{3pt}
\begin{tabular}{@{}l l l r r@{}}
\hline
\textbf{Organ} & \textbf{Modality} & \textbf{Dataset} & \textbf{Pos.} & \textbf{Neg.} \\
\hline
\multirow{4}{*}{Brain} & \multirow{4}{*}{MRI}
 & ADNI-1~\cite{jack2008alzheimer}          & 200   & 221    \\
 & & ADNI-2~\cite{jack2008alzheimer}        & 159   & 232    \\
 & & ADNI-3~\cite{jack2008alzheimer}        & 85    & 431    \\
 & & AIBL~\cite{ellis2009australian}        & 78    & 477    \\
\midrule
\multirow{4}{*}{Chest} & \multirow{4}{*}{X-Ray}
 & RSNA~\cite{wang2017chestx}               & 6,012 & 20,672 \\
 & & Child CXR~\cite{kermany2018identifying}& 4,273 & 1,583  \\
 & & LDD~\cite{ldd}                         & 5,776 & 3,919  \\
 & & CRD~\cite{crd}                         & 9,237 & 10,319 \\
\midrule
\multirow{2}{*}{Eye} & SLO
 & \multirow{2}{*}{FairDomain~\cite{tian2024fairdomain}} & 4,453 & 5,547 \\
 & OCT & & 4,453 & 5,547 \\
\hline
\end{tabular}
\end{minipage}
\hfill
\begin{minipage}[t]{0.57\linewidth}
\centering
\textbf{(b) UDA Algorithms}\\[0.01cm]
\begin{tabularx}{\linewidth}{@{}>{\raggedright\arraybackslash}X >{\raggedright\arraybackslash}X@{}}
\hline
Feat. Dist.: MMD~\cite{long2015learning} & Pseudo Lab.: ATDOC~\cite{liang2021domain} \\
Info. Max: MCC~\cite{jin2020minimum} & Cls. Disc.: MCD~\cite{saito2018maximum} \\
SVD Loss: BNM~\cite{cui2020towards} & Adv. Align: DANN~\cite{ganin2016domain}, CDAN~\cite{long2018conditional}, DALN~\cite{chen2022reusing} \\
\multicolumn{2}{@{}l@{}}{Medical-Spec.: AD2A (Brain MRI)~\cite{guan2021multi}, CoUDA (CXR)~\cite{zhang2020collaborative}} \\
\hline
\end{tabularx}

\vspace{0.96cm}

\textbf{(c) Validators}\\[0.01cm]
\begin{tabularx}{\linewidth}{@{}>{\raggedright\arraybackslash}p{2.1cm} >{\raggedright\arraybackslash}X@{}}
\hline
Source-Guided:  & Source-Risk~\cite{ganin2015unsupervised}, IWCV~\cite{sugiyama2007covariate}, DEV~\cite{you2019towards}, DEV-N~\cite{musgrave2022three} \\
Target-Based:   & Entropy~\cite{morerio2017minimal}, InfoMax~\cite{musgrave2021unsupervised}, Corr-C~\cite{tu2024does}, BNM (V)~\cite{musgrave2022three}, MCC (V)~\cite{jin2020minimum}, SND~\cite{saito2021tune}, ClassAMI~\cite{musgrave2022three}, MixVal~\cite{hu2023mixed}, TransScore~\cite{yang2024can} \\
\hline
\end{tabularx}
\end{minipage}
\end{table*}

We study the complete UDA pipeline in medical imaging under the conditions a clinical deployment would face, focusing on classification tasks such as disease diagnosis. The pipeline comprises two stages. In the adaptation stage, UDA algorithms share a common structure, a backbone, a classification head, and an adaptation module, and optimize a joint objective
\begin{equation}
\mathcal{L} = \mathcal{L}_{\text{cls}} + \lambda\,\mathcal{L}_{\text{adapt}},
\label{eq:uda_objective}
\end{equation}
where $\mathcal{L}_{\text{cls}}$ supervises the model with source labels, $\mathcal{L}_{\text{adapt}}$ aligns feature distributions across domains, and $\lambda$ controls the adaptation strength. Running an algorithm with a given $\lambda$ produces a training run, during which we save a sequence of checkpoints. Each checkpoint $\theta$ is thus uniquely defined by its algorithm, its adaptation strength $\lambda$, and its training iteration, and every checkpoint is a candidate for deployment. 


In the selection stage, we replicate the constraint of clinical deployment: target labels are unavailable, so the model to deploy must be chosen without them. The role of a validator is to provide a label-free validation score for the target performance. Formally, a validator is a scoring function
\begin{equation}
V : \theta \mapsto s \in \mathbb{R},
\label{eq:validator}
\end{equation}
that maps a checkpoint $\theta$ to a scalar score $s = V(\theta)$ without using target labels. The
selected model for deployment is the checkpoint with the best validation score,
\begin{equation}
\theta^{\star} = \arg\max_{\theta \in \Theta} V(\theta),
\label{eq:selection}
\end{equation}
where $\Theta$ is the candidate checkpoint pool and scores are oriented so that higher is better.

\noindent\textbf{Datasets.} To construct medical UDA scenarios, four widely adopted brain MRI datasets are used: ADNI-1, ADNI-2, ADNI-3~\cite{jack2008alzheimer}, and AIBL~\cite{ellis2009australian}. Subjects appearing in multiple ADNI datasets are kept in only one to prevent data leakage. In addition, four publicly available CXR datasets are used: RSNA~\cite{wang2017chestx}, Child CXR~\cite{kermany2018identifying}, LDD~\cite{ldd}, and CRD~\cite{crd}. For retinal imaging, we use FairDomain dataset~\cite{tian2024fairdomain}, which provides paired SLO and OCT acquisitions of the same cohort. For brain MRI and CXR, each dataset is treated as a separate domain, and transfer is
performed across datasets within each modality; for retinal data, transfer is performed between SLO and OCT modalities. Brain MRI datasets contain Alzheimer's disease and cognitively normal subjects, CXR datasets contain pneumonia and non-pneumonia subjects, and retinal datasets contain glaucoma and non-glaucoma subjects. All modalities undergo standard preprocessing following prior work~\cite{guan2021multi,zhang2020collaborative,tian2024fairdomain}. Dataset statistics are summarized in Table~\ref{tab:benchmark_overview} (a).

\noindent\textbf{Methods.} Since we study the complete pipeline as a whole, we include a representative set of
UDA algorithms and, crucially, a diverse set of label-free validators. As shown in Table~\ref{tab:benchmark_overview}~(b) and~(c), the ten algorithms span multiple paradigms, including feature-distance minimization~\cite{long2015learning}, adversarial alignment~\cite{ganin2016domain,long2018conditional,chen2022reusing}, information maximization~\cite{jin2020minimum}, SVD loss~\cite{cui2020towards}, pseudo-labeling~\cite{liang2021domain}, classifier discrepancy~\cite{saito2018maximum}, and medical-specific techniques~\cite{guan2021multi,zhang2020collaborative}. The 13
validators cover both source-guided criteria~\cite{ganin2015unsupervised,musgrave2022three,sugiyama2007covariate,you2019towards} and target-based ones~\cite{morerio2017minimal,musgrave2021unsupervised,tu2024does,jin2020minimum,saito2021tune,yang2024can,hu2023mixed}.

\noindent\textbf{Experimental Setup.} Following Guan et al.~\cite{guan2021multi}, five UDA scenarios (source$\rightarrow$target) are constructed for brain MRI:
ADNI-1$\rightarrow$ADNI-2, ADNI-1$\rightarrow$ADNI-3, ADNI-2$\rightarrow$ADNI-1, ADNI-2$\rightarrow$ADNI-3, and ADNI-1+2$\rightarrow$AIBL. Following Feng et al.~\cite{feng2023contrastive} and Liu et al.~\cite{liu2023attention}, two CXR transfers are used, each in both directions: RSNA$\leftrightarrow$Child CXR and LDD$\leftrightarrow$CRD. For retinal imaging, the two cross-modality directions SLO$\rightarrow$OCT and OCT$\rightarrow$SLO are evaluated. Both source and target data are split into training and validation sets, and target performance is
measured on the target validation set using balanced accuracy.  For datasets without predefined splits, stratified five-fold cross-validation is performed; for datasets with explicit train/test splits, results are averaged over three random seeds. In both cases the mean $\pm$ standard deviation is reported.

\begin{table*}[h]
\centering
\caption{Complete UDA results on ADNI-1$\rightarrow$ADNI-2 (target accuracy, \%; eight best validators by \textbf{Across-Algo} accuracy shown). Each cell reports the target accuracy of the checkpoint a validator selects for an algorithm. The per-algorithm columns report selection within a single algorithm, \textbf{Avg.} is their mean, and \textbf{Across-Algo} pools the checkpoints of all algorithms and selects across them. \textit{Oracle} selects using target labels and is the actual best model, \textit{SourceOnly} is trained on source data only, and \textit{TargetOnly} is trained on labeled target data.}
\label{tab:main_result}
\footnotesize
\setlength{\tabcolsep}{3pt}
\renewcommand{\arraystretch}{1.1}
\resizebox{\textwidth}{!}{%
\begin{tabular}{l | c | *{9}{c} : c | c | c}
\hline
 & \textit{SourceOnly} & {MMD} & {DANN} & {CDAN} & {DALN} & {MCC} & {BNM} & {ATDOC} & {MCD} & {AD2A} & \textbf{Avg.} & \textbf{Across-Algo} & \textit{TargetOnly} \\
\hline
\textit{Oracle} & 88.2\tiny$\pm$2.0 & 88.0\tiny$\pm$3.3 & 90.0\tiny$\pm$4.0 & 89.9\tiny$\pm$3.0 & 90.5\tiny$\pm$6.8 & 89.9\tiny$\pm$1.3 & 89.8\tiny$\pm$4.0 & 89.5\tiny$\pm$1.5 & 87.4\tiny$\pm$4.3 & 92.1\tiny$\pm$0.93 & 89.7\tiny$\pm$1.3 & 93.2\tiny$\pm$2.0 & 90.9\tiny$\pm$2.4 \\
\hline
DEV-N & 80.9\tiny$\pm$5.8 & 82.4\tiny$\pm$2.5 & 84.8\tiny$\pm$2.0 & 81.8\tiny$\pm$8.4 & 84.9\tiny$\pm$8.7 & 81.8\tiny$\pm$2.8 & 81.5\tiny$\pm$0.82 & 79.0\tiny$\pm$6.2 & 81.0\tiny$\pm$6.2 & 87.6\tiny$\pm$1.9 & - & 85.7\tiny$\pm$1.2 & - \\
Source-Risk & 81.2\tiny$\pm$5.8 & 82.8\tiny$\pm$3.1 & 81.8\tiny$\pm$4.0 & 81.1\tiny$\pm$8.0 & 85.9\tiny$\pm$7.4 & 81.6\tiny$\pm$2.5 & 82.9\tiny$\pm$3.7 & 79.0\tiny$\pm$6.2 & 81.5\tiny$\pm$6.3 & 86.8\tiny$\pm$2.8 & - & 84.4\tiny$\pm$2.5 & - \\
DEV & 78.0\tiny$\pm$11 & 81.5\tiny$\pm$3.2 & 84.5\tiny$\pm$6.2 & 82.0\tiny$\pm$6.7 & 84.0\tiny$\pm$10 & 70.8\tiny$\pm$9.6 & 74.0\tiny$\pm$6.9 & 71.2\tiny$\pm$21 & 68.6\tiny$\pm$19 & 81.4\tiny$\pm$8.2 & - & 84.4\tiny$\pm$5.9 & - \\
InfoMax & 84.2\tiny$\pm$2.1 & 82.4\tiny$\pm$2.1 & 87.3\tiny$\pm$4.4 & 86.0\tiny$\pm$1.5 & 89.5\tiny$\pm$6.7 & 76.2\tiny$\pm$8.1 & 84.6\tiny$\pm$7.4 & 80.2\tiny$\pm$11 & 83.3\tiny$\pm$5.6 & 86.3\tiny$\pm$2.2 & - & 83.2\tiny$\pm$7.2 & - \\
BNM (V) & 84.7\tiny$\pm$2.1 & 82.4\tiny$\pm$2.1 & 87.5\tiny$\pm$4.0 & 84.9\tiny$\pm$3.8 & 87.2\tiny$\pm$3.4 & 76.2\tiny$\pm$8.1 & 84.6\tiny$\pm$7.4 & 80.2\tiny$\pm$11 & 83.3\tiny$\pm$5.6 & 86.3\tiny$\pm$2.2 & - & 81.8\tiny$\pm$9.6 & - \\
ClassAMI & 80.9\tiny$\pm$6.4 & 58.1\tiny$\pm$6.8 & 82.1\tiny$\pm$3.1 & 82.7\tiny$\pm$3.8 & 84.0\tiny$\pm$6.4 & 82.0\tiny$\pm$1.5 & 80.0\tiny$\pm$2.3 & 85.1\tiny$\pm$1.1 & 59.2\tiny$\pm$9.8 & 81.6\tiny$\pm$2.1 & - & 81.0\tiny$\pm$2.2 & - \\
SND & 73.9\tiny$\pm$12 & 84.2\tiny$\pm$4.2 & 80.0\tiny$\pm$2.6 & 79.2\tiny$\pm$5.8 & 81.8\tiny$\pm$3.9 & 77.6\tiny$\pm$2.3 & 77.3\tiny$\pm$9.4 & 83.6\tiny$\pm$2.7 & 82.2\tiny$\pm$5.6 & 80.6\tiny$\pm$5.3 & - & 80.7\tiny$\pm$8.7 & - \\
MCC (V) & 84.2\tiny$\pm$2.7 & 81.9\tiny$\pm$3.6 & 84.9\tiny$\pm$3.2 & 84.5\tiny$\pm$4.0 & 85.0\tiny$\pm$6.5 & 76.2\tiny$\pm$8.1 & 87.1\tiny$\pm$2.7 & 87.9\tiny$\pm$3.8 & 83.1\tiny$\pm$3.7 & 87.1\tiny$\pm$3.6 & - & 80.4\tiny$\pm$8.3 & - \\

\hline
\end{tabular}%
}
\end{table*}

\noindent\textbf{Implementation Details.} For all experiments, the classification head is a three-layer MLP with a dropout rate of 0.5. For brain MRI, a 3D ResNet-50~\cite{he2016deep} trained from scratch is used as the backbone, with a batch size of eight per domain. For CXR and retinal imaging, a 2D ResNet-50 pretrained on ImageNet~\cite{deng2009imagenet} is used, with a batch size of 48 per domain. All algorithms are trained with AdamW (weight decay 1e-4) and a one-cycle learning rate schedule with warm-up and a peak learning rate of 1e-3, for 10k iterations on brain MRI and retinal imaging and 30k iterations on CXR. The adaptation strength is varied over $\lambda\in\{0.1, 0.5, 1.0\}$, giving three runs per algorithm. After warm-up, checkpoints are saved at uniform intervals, yielding 50 checkpoints per run and 150 checkpoints per algorithm. $\mathcal{L}_{\text{cls}}$ is binary cross-entropy with class-balanced weighting derived from the source labels. All runs are conducted on an A6000 GPU.

\section{Results and Discussion}

We use ADNI-1$\rightarrow$ADNI-2 as an illustrative example for the complete UDA pipeline in Table~\ref{tab:main_result}, which reports the target accuracy of the checkpoint each validator selects. This selection is compared against the \textit{Oracle}, which selects directly with target labels. We define the \textit{selection gap} as the difference in target accuracy between the \textit{Oracle}-selected model and the model selected by a label-free validator. The per-algorithm columns report selection within a single algorithm, and \textbf{Avg.} is their mean, while \textbf{Across-Algo} pools all algorithms' checkpoints and selects across them. Due to space constraints, we display the eight best validators by their \textbf{Across-Algo} accuracy. Full results for all scenarios can be found in the supplementary material. We report \textit{SourceOnly} (no adaptation) and \textit{TargetOnly} (trained on labeled target data) as references.

\subsection{Adaptation Works in Principle}

\begin{figure*}
    \centering
    \includegraphics[width=1.0\linewidth]{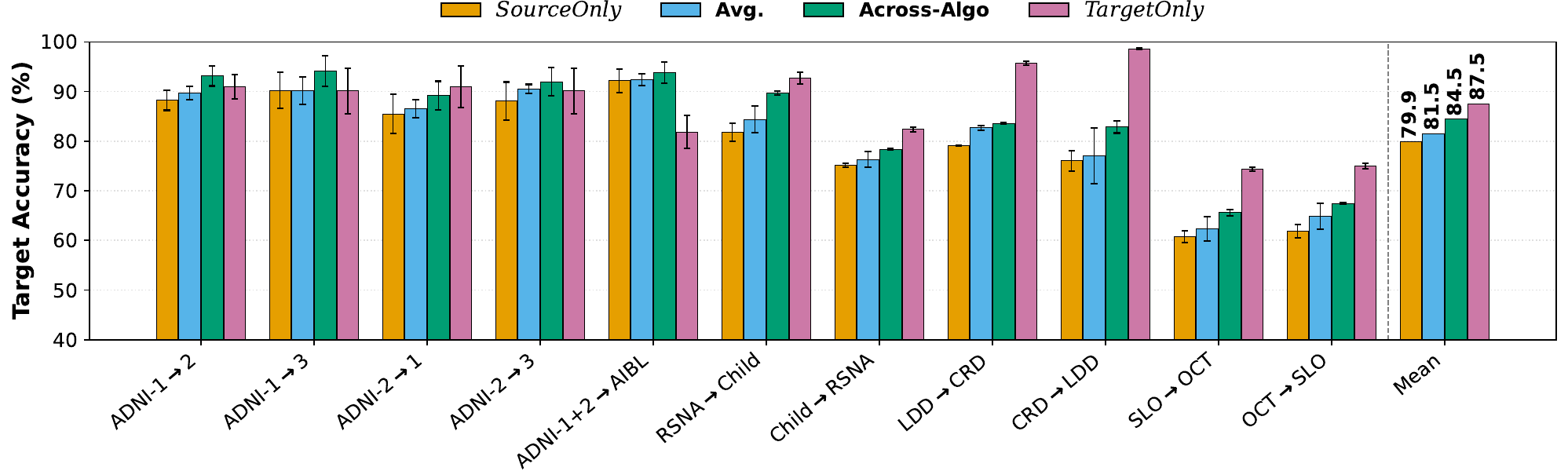}
    \caption{Target accuracy under \textit{Oracle} selection across UDA scenarios. Bars show \textit{SourceOnly} (no adaptation), the per-algorithm average (\textbf{Avg.}), \textbf{Across-Algo} (selection over all algorithms' pooled checkpoints), and \textit{TargetOnly} (labeled-target). \textbf{Across-Algo} exceeds \textit{SourceOnly} in all scenarios and approaches \textit{TargetOnly}, indicating a capable adapted model usually exists. Error bars: standard deviation.}
\label{fig:adaptation_effective}
\end{figure*}
As shown in Table~\ref{tab:main_result}, under \textit{Oracle} selection, ADNI-1$\rightarrow$ADNI-2 scenario reaches 93.2\% accuracy when pooling all checkpoints (\textbf{Across-Algo}), exceeding both \textit{SourceOnly} (88.2\%) and
\textit{TargetOnly} (90.9\%). The gain is broad: the \textit{Oracle}-selected checkpoint beats \textit{SourceOnly} for seven of the nine algorithms, with a per-algorithm average of 89.7\% (\textbf{Avg.}), and surpasses \textit{TargetOnly} for
one.

Figure~\ref{fig:adaptation_effective} extends Table~\ref{tab:main_result} to all eleven UDA scenarios, reporting the same four quantities under \textit{Oracle} selection. At the per-algorithm level, the average (\textbf{Avg.}) exceeds \textit{SourceOnly} in ten of the eleven scenarios, and in three it even matches or surpasses \textit{TargetOnly}. Pooling checkpoints across algorithms strengthens this further: \textbf{Across-Algo} lies above \textit{SourceOnly} in every scenario, and exceeds \textit{TargetOnly} in four. Averaged over all scenarios (\textbf{Mean}), the ordering is consistent: \textit{SourceOnly} $<$ \textbf{Avg.} $<$ \textbf{Across-Algo}, with \textbf{Across-Algo} reaching 84.5\% and approaching \textit{TargetOnly} (87.5\%). Adaptation is therefore often effective across these clinical UDA scenarios; in particular, a capable adapted model exists in most scenarios (evidenced by \textbf{Across-Algo}). Nevertheless, this model is only useful if it can be identified without target labels, as deployment requires.

\subsection{Large Selection Gap from Validators}
\begin{figure*}
    \centering
    \includegraphics[width=1.0\linewidth]{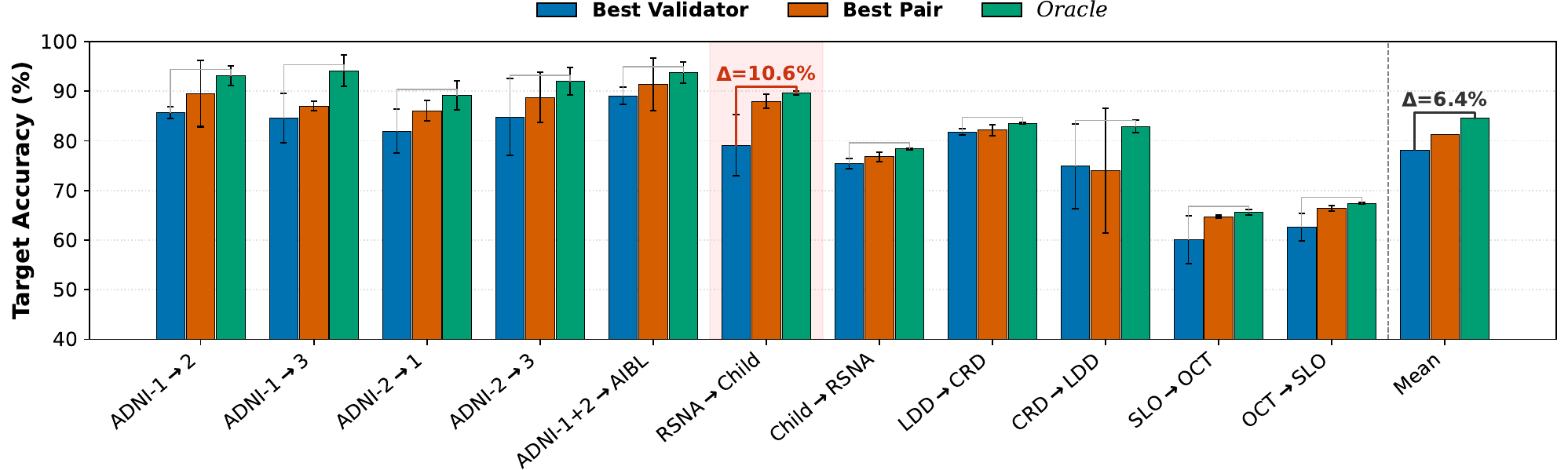}
    \caption{Target accuracy across UDA scenarios under three selections (\%). Bars show the accuracy of the model selected by the best validator over the across-algorithm pool (\textbf{Best Validator}), by the best algorithm--validator combination (\textbf{Best Pair}), and by the across-algorithm \textit{Oracle}. The first two fall below the \textit{Oracle} in every scenario, with the gap up to 10.6 points and averaging 6.4 (\textbf{Mean}). Note that \textbf{Best Validator} and \textbf{Best Pair} differ from scenario to scenario and are identified using target labels, so they cannot be chosen in advance; the gap realized at deployment can only be larger. Error bars: standard deviation.}
\label{fig:selection_gap}
\end{figure*}
The capable model established above can only be deployed if a validator selects it. Return to ADNI-1$\rightarrow$ADNI-2 scenario as shown in Table~\ref{tab:main_result}, under \textbf{Across-Algo}, the \textit{Oracle} attains 93.2\%. Over the same across-algorithm pool, the model selected by the best validator (DEV-N) reaches 85.7\%, and that selected by the best algorithm--validator pair (DALN with InfoMax) reaches 89.5\%. Even these best cases fall 3.7 to 7.5 points short of the \textit{Oracle}.

Figure~\ref{fig:selection_gap} extends Table~\ref{tab:main_result} to all eleven UDA
scenarios, reporting the target accuracy of the model selected by the best validator over the across-algorithm pool (\textbf{Best Validator}), by the best algorithm--validator pair (\textbf{Best Pair}), and by the across-algorithm \textit{Oracle}. In every scenario, both \textbf{Best Validator} and \textbf{Best
Pair} fall below the \textit{Oracle}: the gap reaches up to 10.6 points (RSNA$\rightarrow$Child CXR) and averages 6.4 points across scenarios (\textbf{Mean}).
Crucially, \textbf{Best Validator} and \textbf{Best Pair} differ from scenario to scenario and are themselves identified using target labels; at deployment, where no labels are available to choose them in advance, the realized gap can only be larger.
A capable model exists in the pool, but selecting it without labels leaves a substantial portion of the achievable performance unrealized.

\noindent\textbf{The Selection Gap Persists Across Architectures.} To test whether our findings depend on the backbone, we repeat RSNA$\rightarrow$Child CXR with three additional architectures spanning distinct families: ConvNeXt~\cite{liu2022convnet} (a modern CNN), ResMLP~\cite{touvron2022resmlp} (an MLP-based model), and DeiT~\cite{touvron2021training} (a vision transformer). As shown in Table~\ref{tab:backbones}, the model selected by the best validator leaves a noticeable gap to the \textit{Oracle} for every architecture. This suggests the selection gap is not specific to a particular backbone.

\begin{figure}[t]
\centering
\begin{subfigure}{0.49\linewidth}
  \includegraphics[width=\linewidth]{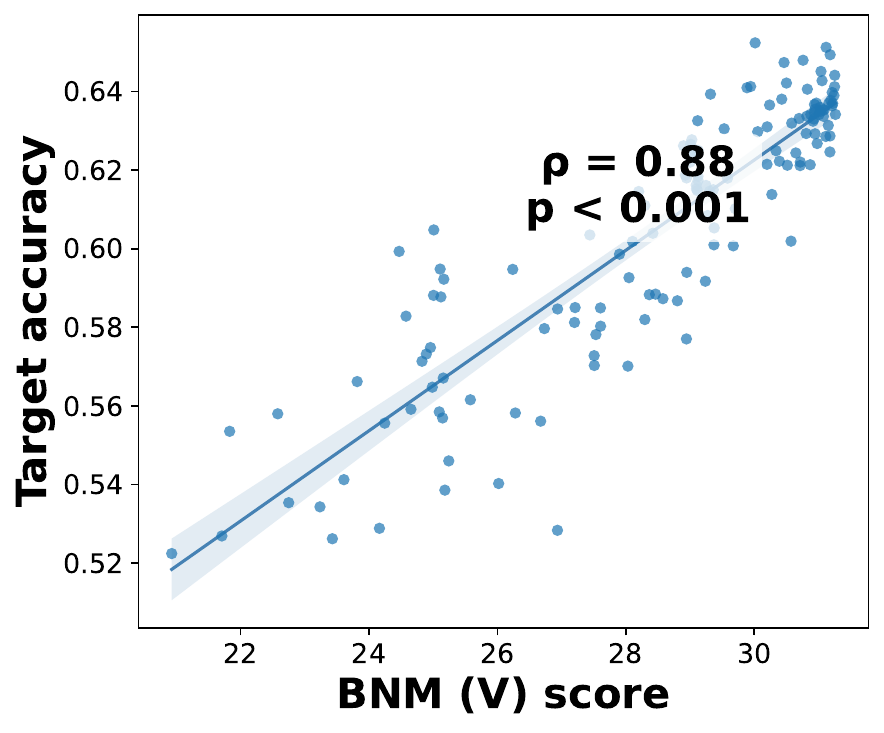}
  \caption{OCT$\rightarrow$SLO, DALN}
  \label{fig:pair_a}
\end{subfigure}
\hfill
\begin{subfigure}{0.49\linewidth}
  \includegraphics[width=\linewidth]{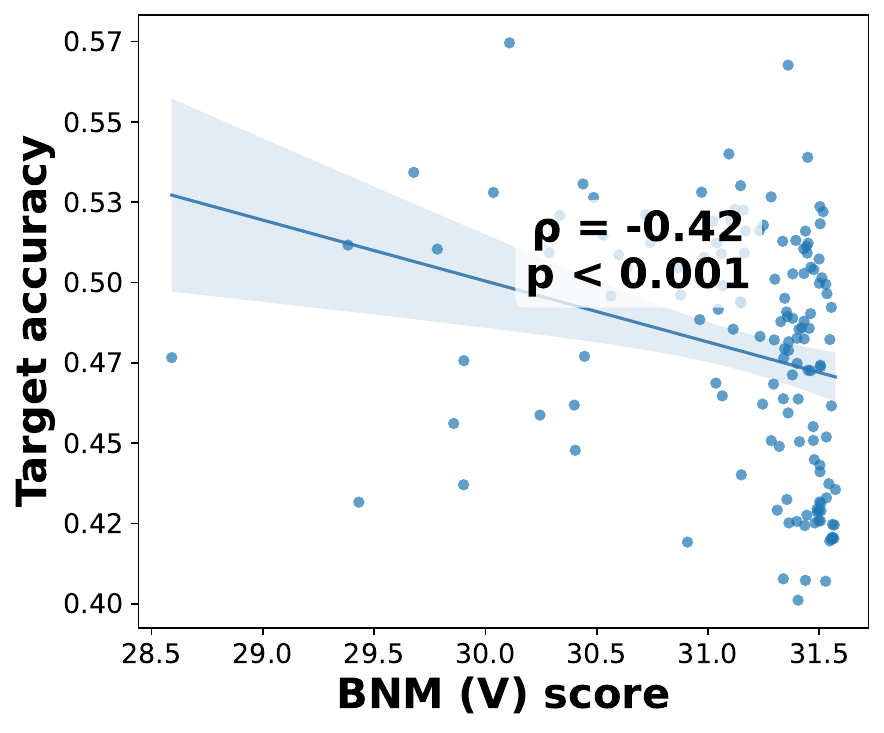}
  \caption{OCT$\rightarrow$SLO, ATDOC}
  \label{fig:pair_b}
\end{subfigure}

\vspace{0.5em}

\begin{subfigure}{0.49\linewidth}
  \includegraphics[width=\linewidth]{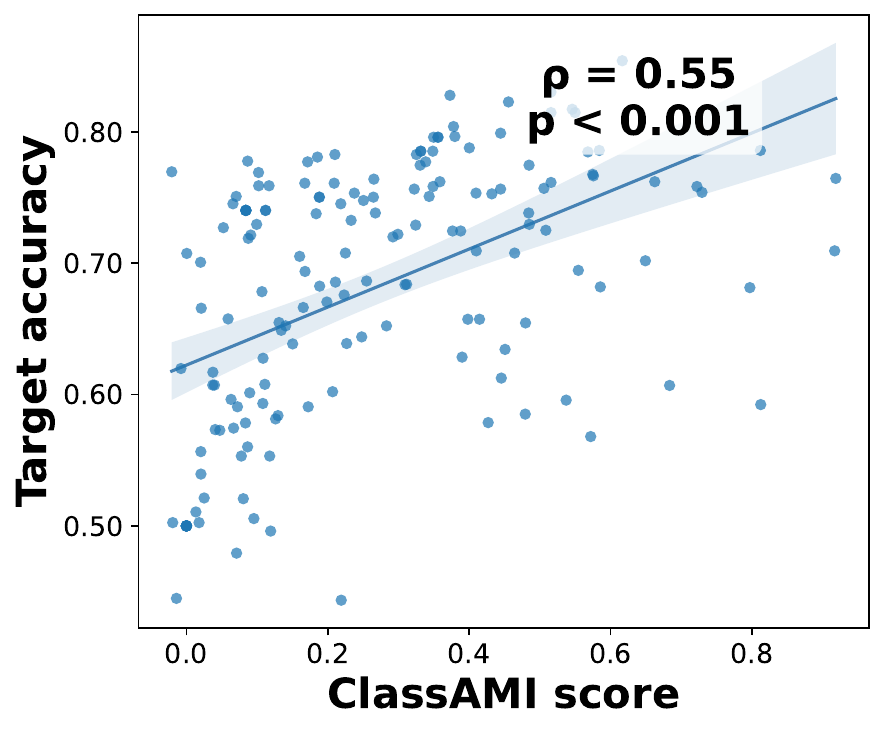}
  \caption{MMD, ADNI-2$\rightarrow$ADNI-1}
  \label{fig:pair_c}
\end{subfigure}
\hfill
\begin{subfigure}{0.49\linewidth}
  \includegraphics[width=\linewidth]{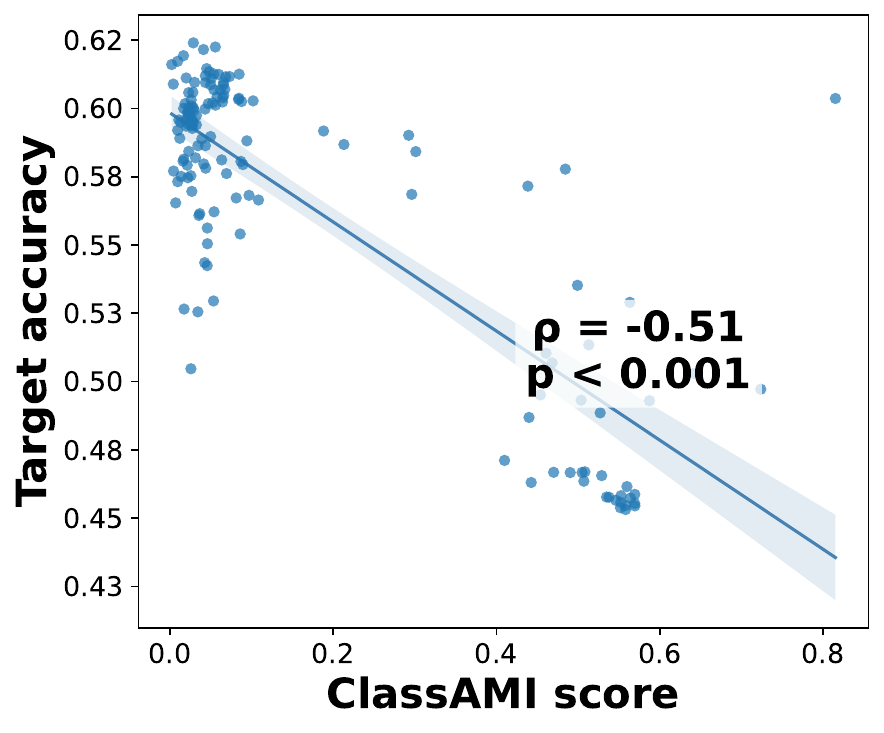}
  \caption{MMD, SLO$\rightarrow$OCT}
  \label{fig:pair_d}
\end{subfigure}
\caption{Within-algorithm Spearman correlation ($\rho$) between validation score and target accuracy. Each point is a checkpoint. The same validator can flip from reliable to reversed with a change of algorithm (a, b) or scenario (c, d) alone.}
\label{fig:spearman_pairs}
\end{figure}

\begin{table}[t]
\centering
\caption{Target accuracy (\%) on RSNA$\rightarrow$Child CXR for four backbones. \textbf{Best Val.} is the accuracy of the model selected by the best validator under the across-algorithm pool, and \textit{Oracle} that of the actual best model. The gap persists across all four backbones.}
\label{tab:backbones}
\scriptsize
\setlength{\tabcolsep}{5pt}
\renewcommand{\arraystretch}{1.1}
\begin{tabular}{l| c c c c}
\hline
\textbf{Backbone} & \textbf{Best Val.} & \textit{Oracle} & $\Delta$\textit{Oracle} \\
\hline
ResNet-50   & 79.1 & 89.7 & 10.6 \\
ResMLP      & 70.9 & 79.2 & 8.3 \\
ConvNeXt    & 82.6 & 90.2 & 7.6 \\
DeiT        & 80.4 & 86.5 & 6.1 \\
\hline
\end{tabular}
\end{table}

\subsection{Structural Selection Gap from Validators}
Having established the magnitude of the selection gap, we now investigate its origin. The gap may arise either from a suboptimal choice of validator, which a more suitable one would remedy, or from a more structural limitation that no evaluated validator overcomes. Since selection is fundamentally a ranking problem (a validator succeeds by ranking checkpoints so that the best one scores highest), we trace the selection gap to validator reliability, measured by the Spearman rank correlation $\rho$ between validation scores and true target accuracy. For validators where a lower score indicates a better checkpoint, the sign is flipped so that a positive correlation always denotes the intended direction.

\noindent\textbf{Within-Algorithm Selection.} Figure~\ref{fig:spearman_pairs}~(a) and (b) fix the scenario OCT$\rightarrow$SLO and validator BNM (V) and vary the algorithm: BNM (V) ranks DALN's checkpoints well and in the designed direction ($\rho=0.88, p < 0.001$), yet ranks ATDOC's checkpoints in reverse ($\rho=-0.42, p<0.001$). Figure~\ref{fig:spearman_pairs}~(c) and (d) fix the algorithm (MMD) and validator (ClassAMI) and vary the scenario: the correlation is positive as intended for ADNI-2$\rightarrow$ADNI-1 ($\rho=0.53,p<0.001$) but reversed for SLO$\rightarrow$OCT ($\rho=-0.51,p<0.001$). Further individual analyses can be found in the supplementary material. Figure~\ref{fig:local_spearman_heatmaps} aggregates the within-algorithm Spearman correlation between validation scores and target accuracy two ways: (a) averaged over algorithms for a per-scenario view, and (b) averaged over scenarios for a per-algorithm view; in both, blue denotes correlation in the intended direction (darker is stronger) and warm red the opposite. On average, validators are moderately informative, but their reliability is uneven: no validator attains consistent correlation across all scenarios in~(a) or all algorithms in~(b), and the strongest validator differs from column to column. Source-Risk, for example, is the most reliable validator for ADNI-2$\rightarrow$ADNI-3 ($\rho=0.68$) yet drops to near zero for Child CXR$\rightarrow$RSNA ($\rho=0.07$); likewise, InfoMax is strongest for AD2A ($\rho=0.82$) but falls to $\rho=0.20$ for ATDOC.

\begin{figure*}[t]
    \centering
    \begin{subfigure}{0.49\linewidth}
        \includegraphics[width=\linewidth]{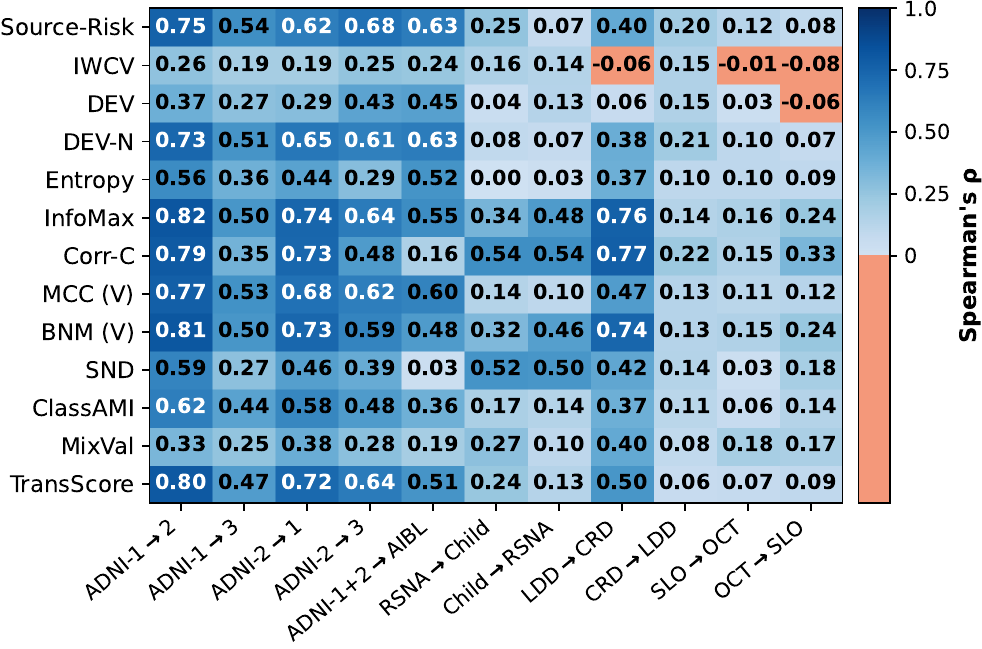}
        \caption{Per-scenario view (averaged over algorithms)}
        \label{fig:spearman_local_scenario}
    \end{subfigure}
    \hfill
    \begin{subfigure}{0.49\linewidth}
        \includegraphics[width=\linewidth]{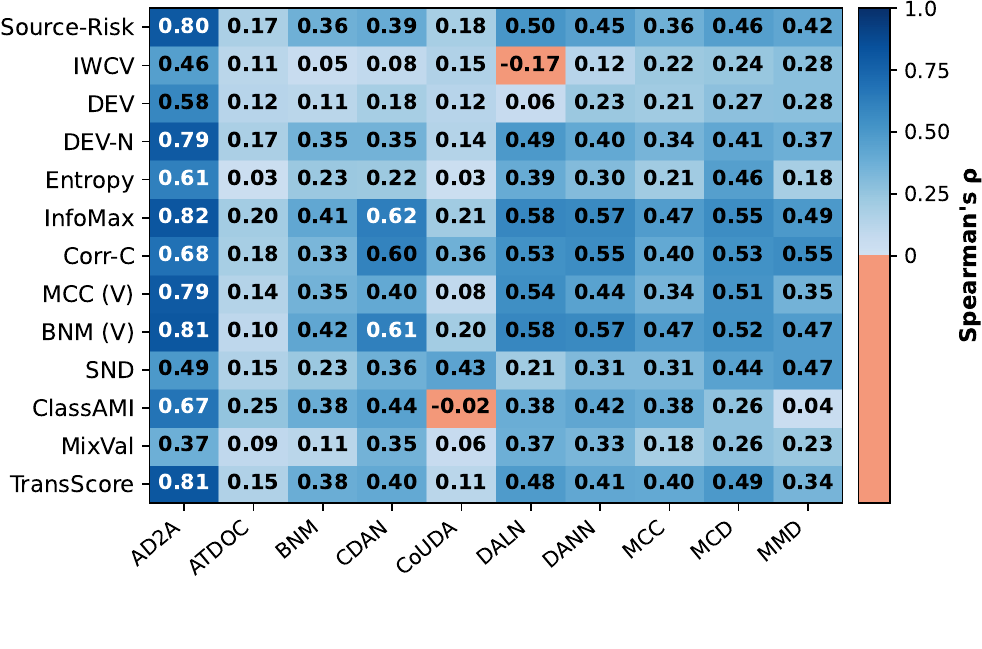}
        \caption{Per-algorithm view (averaged over scenarios)}
        \label{fig:spearman_local_algo}
    \end{subfigure}
   \caption{Within-algorithm Spearman correlation ($\rho$) between validation scores and target accuracy. Blue denotes correlation in the intended direction (darker is stronger), warm red the opposite. No validator is consistently reliable across all scenarios or algorithms, and the most reliable one differs from column to column.}
    \label{fig:local_spearman_heatmaps}
\end{figure*}

\begin{figure}[t]
\centering
\begin{subfigure}{0.49\linewidth}
  \includegraphics[width=\linewidth]{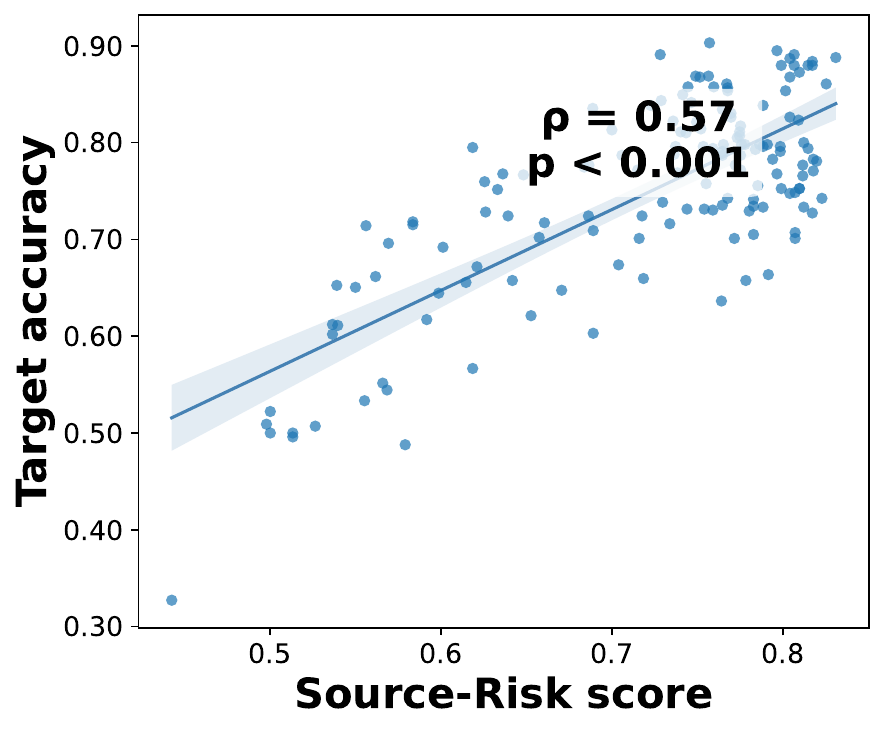}
  \caption{Within BNM algorithm}
  \label{fig:global_a}
\end{subfigure}
\hfill
\begin{subfigure}{0.49\linewidth}
  \includegraphics[width=\linewidth]{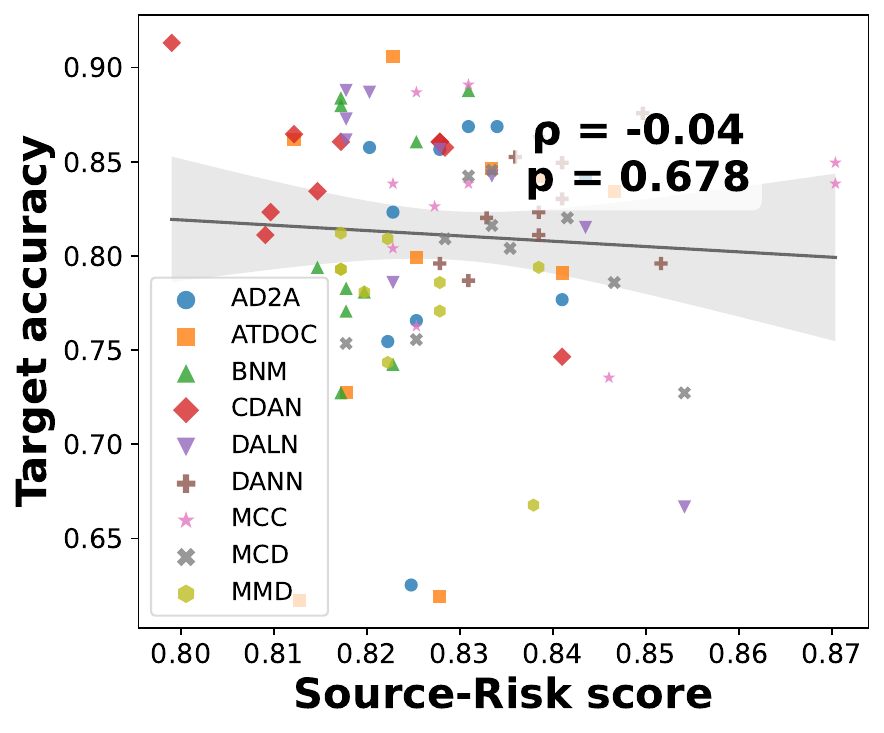}
  \caption{Across Algorithms}
  \label{fig:global_b}
\end{subfigure}
\caption{Spearman correlation ($\rho$) between Source-Risk scores and target accuracy on ADNI-1$\rightarrow$ADNI-2. Each point is a checkpoint. {(a)} Within BNM algorithm, the correlation is moderate. {(b)} Across all algorithms, it collapses to near zero.}
\label{fig:spearman_global}
\end{figure}

\begin{figure}[t]
\centering
\includegraphics[width=\linewidth]{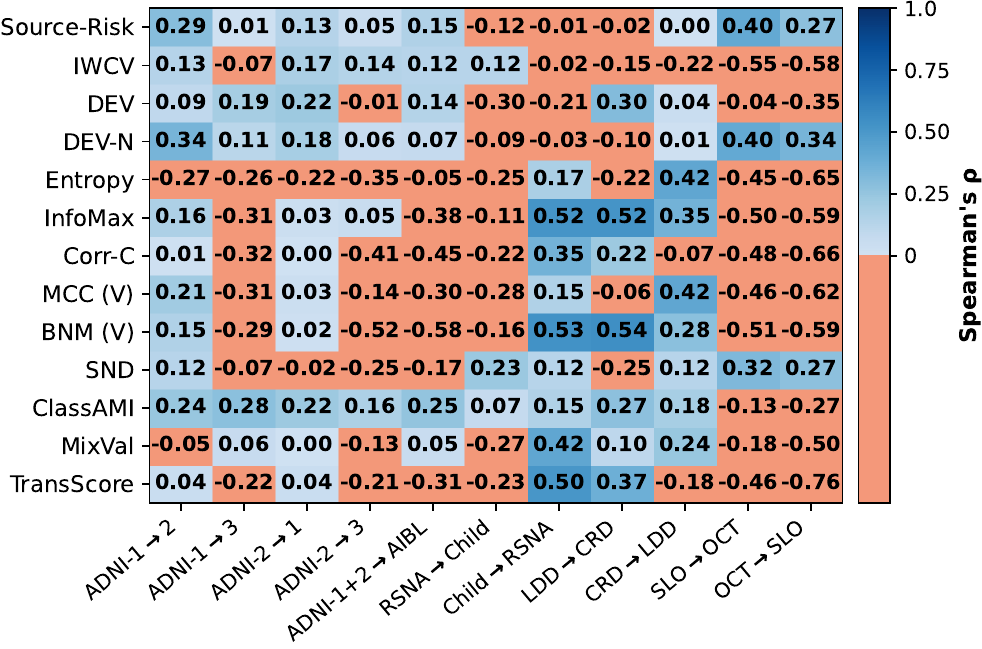}
\caption{Across-algorithm Spearman correlation ($\rho$) between validation scores and target accuracy. Blue denotes correlation in the intended direction (darker is stronger), warm red the opposite. Most cells are weak or negative, showing that most validators are unreliable in this across-algorithm setting.}
\label{fig:global_spearman_heatmaps}
\end{figure}

\noindent\textbf{Across-Algorithm Selection.} Figure~\ref{fig:spearman_global}~(a) shows Source-Risk on ADNI-1$\rightarrow$ADNI-2 for a single algorithm (BNM): within this algorithm, it ranks checkpoints reasonably well ($\rho=0.57,p<0.001$). Figure~\ref{fig:spearman_global}~(b) instead pools checkpoints across all algorithms for the same scenario and validator. To form this pool while keeping the comparison meaningful, we take, for each algorithm, the ten checkpoints that the validator scores highest: this retains the competitive checkpoints worth selecting among, while ten per algorithm provides a sufficient sample to estimate the rank correlation. On this pooled set, the correlation collapses to near zero ($\rho=-0.04,p=0.678$). Figure~\ref{fig:global_spearman_heatmaps} reports the same across-algorithm correlation across all validators and scenarios: most validators are unreliable or contradict their intended direction, with correlations weak or negative.

\noindent For within-algorithm selection, a suitable validator may exist among evaluated ones, but which one differs by algorithm and scenario, so it cannot be chosen in advance. For across-algorithm selection, even this fails: most evaluated validators correlate weakly or in reverse. The gap therefore appears largely structural rather than a matter of choosing a better validator among those evaluated ones.

\subsection{Strategies to Narrow the Selection Gap}
\label{sec:remedies}

The structural selection gap leaves a large distance between what adaptation achieves and what label-free selection realizes at deployment. In the absence of a validator consistently reliable across scenarios, we explore two practical strategies and show that each recovers part of this gap.

\noindent\textbf{Ensembling.} Since no single evaluated validator is consistently reliable, we avoid committing to one and instead ensemble the checkpoints selected by different validators, similar to Hu et al.~\cite{hu2024towards} but additionally ensembling across algorithms. In the across-algorithm setting, where checkpoints from all algorithms are pooled, we select one checkpoint per algorithm with every validator and ensemble these selections by averaging their predictions. As shown in Table~\ref{tab:ensembling}, ensembling matches or exceeds \textbf{Best Val.} in most scenarios. Again \textbf{Best Val.} is the model selected by the best validator for each scenario, which differs across scenarios and cannot be known in advance. The remaining gap to the \textit{Oracle} is small in several cases (e.g., 1.1 for OCT$\rightarrow$SLO and 2.2 for LDD$\rightarrow$CRD), showing that it recovers much of the achievable performance. It nonetheless remains far from the \textit{Oracle} in other cases such as RSNA$\rightarrow$Child CXR. 


\noindent\textbf{Small Target Labeling Budget.} Since fully label-free selection cannot be trusted under the evaluated validators, we explore whether spending a small budget on target labeling can identify a better checkpoint. We label a small subset of the target validation set and select the checkpoint with the highest accuracy on it. The budget is varied over five levels (B$_1$ to B$_{5}$): 5, 10, \dots, 25 labeled samples for brain MRI, whose target validation set is around 100 samples, and 1\% to 5\% for the others. Each budget is repeated over ten random trials, and we report the mean. As shown in Table~\ref{tab:ensembling}, a small budget improves selection steadily, surpassing both \textbf{Best Val.} (78.2) and ensembling (78.0) at B$_5$ (25 labeled samples for brain MRI, 5\% for the other scenarios) on average, though a gap to the \textit{Oracle} still remains.
\begin{table}[h]
\centering
\caption{Two strategies for narrowing the selection gap (target accuracy, \%). \textbf{Ens.} ensembles the checkpoints selected by all validators across all algorithms. \textbf{Small Labeling Budget} selects the best checkpoint on a small labeled target subset, at five levels (B$_1$ to B$_5$; 5--25 samples for brain MRI, 1--5\% otherwise). On average, ensembling nearly matches \textbf{Best Val.} (the best validator under the across-algorithm pool), and the labeling budget surpasses both at B$_5$, though a gap to the \textit{Oracle} remains for both.}
\label{tab:ensembling}
\scriptsize
\setlength{\tabcolsep}{2.5pt}
\renewcommand{\arraystretch}{1.2}
\begin{tabular}{l c c ccccc c cc}
\hline
& \textbf{Ens.} & & \multicolumn{5}{c}{\textbf{Small Labeling Budget}} & & \textbf{Best Val.} & \textit{Oracle} \\
\cmidrule(lr){2-2} \cmidrule(lr){4-8} \cmidrule(lr){10-11}
\textbf{Scenario} & & & B$_1$ & B$_2$ & B$_3$ & B$_4$ & B$_5$ & & & \\
\hline
ADNI-1$\rightarrow$ADNI-2   & 89.0 & & 78.6 & 83.0 & 84.0 & 86.5 & 87.5 & & 85.7 & 93.2 \\
ADNI-1$\rightarrow$ADNI-3   & 88.2 & & 78.7 & 80.8 & 81.1 & 81.2 & 83.3 & & 84.5 & 94.2 \\
ADNI-2$\rightarrow$ADNI-1   & 84.2 & & 76.8 & 79.0 & 81.1 & 82.2 & 83.3 & & 82.0 & 89.2 \\
ADNI-2$\rightarrow$ADNI-3   & 86.7 & & 79.3 & 79.4 & 84.3 & 82.2 & 83.1 & & 84.8 & 92.0 \\
ADNI-1+2$\rightarrow$AIBL   & 89.9 & & 81.6 & 82.9 & 82.9 & 85.1 & 86.0 & & 89.0 & 93.8 \\
\hline
RSNA$\rightarrow$Child CXR  & 80.0 & & 76.5 & 78.4 & 81.5 & 83.1 & 85.0 & & 79.1 & 89.7 \\
Child CXR$\rightarrow$RSNA  & 74.4 & & 74.3 & 75.7 & 76.5 & 76.9 & 76.9 & & 75.4 & 78.4 \\
LDD$\rightarrow$CRD         & 81.3 & & 80.7 & 81.6 & 82.1 & 82.3 & 82.2 & & 81.8 & 83.5 \\
CRD$\rightarrow$LDD         & 55.1 & & 77.1 & 78.5 & 80.5 & 80.4 & 80.5 & & 74.9 & 82.9 \\
\hline
SLO$\rightarrow$OCT         & 63.2 & & 59.3 & 61.3 & 60.8 & 61.3 & 62.6 & & 60.1 & 65.6 \\
OCT$\rightarrow$SLO         & 66.3 & & 60.0 & 60.9 & 63.5 & 64.5 & 64.4 & & 62.7 & 67.4 \\
\hline
\textbf{Mean}               & 78.0 & & 74.8 & 76.5 & 78.0 & 78.7 & 79.5 & & 78.2 & 84.5 \\
\hline
\end{tabular}
\end{table}
\subsection{Discussion}
Our results show that the main barrier between current UDA and clinical deployment lies less in producing a capable adapted model, which usually exists, than in selecting it without target labels. Under \textit{Oracle} selection, such a model exists in the \textbf{Across-Algo} pool in most scenarios, often recovering much of the gap between \textit{SourceOnly} and \textit{TargetOnly} and at times exceeding both, confirming
that adaptation is usually effective. The evaluated validators, however, do not reliably identify this model, leaving
a large and structural gap to the best available one. For within-algorithm selection, a reliable validator may exist, but which one is best shifts from algorithm to algorithm and from scenario to scenario, so no fixed choice can be made in advance. For across-algorithm selection, most validators correlate only weakly with target performance, and many rank checkpoints opposite to their intended direction. Note that we do not optimize \textit{TargetOnly} extensively; in several scenarios it falls below the per-algorithm \textit{Oracle} average (\textbf{Avg.}) or even \textit{SourceOnly}, which may stem from training on the target domain alone, with less labeled data than the combined source and target signal exploited by adaptation, and from target class imbalance (e.g., AIBL).

The unreliability of evaluated validators also explains an observation that might otherwise question the value of adaptation: under many validators, the selected adapted model performs worse than the \textit{SourceOnly} baseline (e.g., MCC with InfoMax in Table~\ref{tab:main_result}). We interpret this as a selection problem, not adaptation alone: a capable adapted model often exists (the \textit{Oracle} confirms it), but an unreliable validator may fail to find it and sometimes picks one worse than not adapting. On the other hand, in Table~\ref{tab:main_result}, under \textit{Oracle} selection MMD and MCD fall below \textit{SourceOnly}, so the limitation here lies also in adaptation. This reflects either our relatively small hyperparameter set ($\lambda$ and the training iteration) or the unsuitability of these algorithms for this transfer. It nonetheless reinforces the need for reliable validators: at the across-algorithm selection scale, a good validator should recognize and avoid such negative transfer.

We also notice that validator reliability, measured by Spearman correlation $\rho$, and the target accuracy of the selected model do not always align. As shown in Figure~\ref{fig:global_spearman_heatmaps} and the \textbf{Across-Algo} column of Table~\ref{tab:main_result}, for ADNI-1$\rightarrow$ADNI-2, ClassAMI has a higher
correlation than InfoMax ($\rho=0.24$ vs.\ $0.16$) yet selects a less accurate model (81.0\% vs.\ 83.2\%). This is because the two measure different things: $\rho$ captures how well a validator orders all checkpoints, whereas selection depends on its single top-scored checkpoint, so a validator with lower $\rho$ can still place a good checkpoint at the top. This suggests two complementary needs: a reliable validator to produce a good ranking, and potentially a further step to identify the truly best one.

\noindent\textbf{How Far from Clinical Deployment?} The answer is two-sided. Under \textit{Oracle} selection, a capable model exists in the checkpoint pool in most scenarios, so adaptation itself is often effective. What is missing is the ability to identify that model without target labels. Deployment therefore depends on the complete pipeline, not adaptation alone: selection remains a largely unsolved step, and closing it would bring much of current UDA closer to clinical use. In some scenarios the achievable accuracy itself remains low, and for some algorithms even the \textit{Oracle}-selected model falls below \textit{SourceOnly}, so better adaptation algorithms are still needed. Yet even there, and for any future algorithm, a capable model still has to be identified. Improving algorithms raises what is achievable, while reliable selection determines what is actually reached at deployment.

The two strategies in Section~\ref{sec:remedies} recover part of this selection gap but do not close it, and both carry practical costs. Ensembling requires training potentially multiple algorithms across their hyperparameters, since which suits a given transfer is not known in advance, and multiple forward passes per image at inference; it is computationally costly and can even harm performance under naive aggregation when poor checkpoints are
included (e.g., CRD$\rightarrow$LDD). The labeling budget, though small in relative terms (5 to 25 volumes for brain MRI, 1\% to 5\% otherwise), carries a real annotation cost: even a single brain MRI volume is time-consuming, and 1\% of a large dataset can amount to hundreds of images (e.g., over 260 for RSNA), more so for richer tasks such as
segmentation or detection. Active selection~\cite{sawade2012active,hara2024active,kay2025consensus}
may match these gains with fewer labels. Overall, reliable label-free selection remains the central open problem for clinically deployable UDA.

\noindent\textbf{Limitations and Future Work.} Our study covers binary classification; extending it to multi-class, multi-label, segmentation, or detection tasks, where the prediction structure differs and selection criteria may need adapting, is a valuable future direction. As classification lies at the core of these more complex tasks, label-free selection may not become easier there, and the gap could even widen. We
evaluate by balanced accuracy, while other metrics such as sensitivity and specificity also matter clinically, and whether validators rank checkpoints well under them remains open.

\section{Conclusion}
In this work, we evaluated the complete UDA pipeline in medical imaging, adaptation and label-free selection together, under clinical deployment conditions. Spanning eleven clinically relevant cross-domain scenarios from nine datasets across brain MRI, CXR, and retinal imaging, ten UDA algorithms, and 13 validators, our evaluation covers over 80{,}000 trained checkpoints. By considering adaptation and selection as a whole, we find that a capable adapted model often exists, but selecting it without target labels is difficult: the validator-selected models leave a large and structural gap to the best available one, as no evaluated validator is consistently reliable, for either within-algorithm or across-algorithm selection. Two strategies, ensembling and a small target-labeling budget, narrow this gap but do not close it entirely, and both have limitations. Overall, deployable UDA depends on both adaptation and selection working together; addressing the less explored selection step could bring much of current UDA closer to clinical use.

\section*{Acknowledgments}
This study was funded by the German Research Foundation
DFG (Project: KEMAI, GRK 3012 -- 520750254) and by
the German Federal Ministry of Research, Technology and
Space BMFTR as part of the University Medicine Network
3.0 (Project: RACOON, 01KX2524).

Data collection and sharing for the Alzheimer's Disease Neuroimaging Initiative (ADNI) is funded by the National Institute on Aging (National Institutes of Health Grant U19 AG024904). The grantee organization is the Northern California Institute for Research and Education. In the past, ADNI has also received funding from the National Institute of Biomedical Imaging and Bioengineering, the Canadian Institutes of Health Research, and private sector contributions through the Foundation for the National Institutes of Health (FNIH) including generous contributions from the following: AbbVie, Alzheimer's Association; Alzheimer's Drug Discovery Foundation; Araclon Biotech; BioClinica, Inc.; Biogen; Bristol-Myers Squibb Company; CereSpir, Inc.; Cogstate; Eisai Inc.; Elan Pharmaceuticals, Inc.; Eli Lilly and Company; EuroImmun; F. Hoffmann-La Roche Ltd and its affiliated company Genentech, Inc.; Fujirebio; GE Healthcare; IXICO Ltd.; Janssen Alzheimer Immunotherapy Research \& Development, LLC.; Johnson \& Johnson Pharmaceutical Research \& Development LLC.; Lumosity; Lundbeck; Merck \& Co., Inc.; Meso Scale Diagnostics, LLC.; NeuroRx Research; Neurotrack Technologies; Novartis Pharmaceuticals Corporation; Pfizer Inc.; Piramal Imaging; Servier; Takeda Pharmaceutical Company; and Transition Therapeutics.

{
    \small
    \bibliographystyle{ieeenat_fullname}
    \bibliography{main}
}

\clearpage
\section{Supplementary Material}

In this supplementary material, we provide the full results of our complete unsupervised domain adaptation (UDA) pipeline for each evaluated clinically relevant cross-domain scenario in Section~\ref{sec:full_results}. In addition, we provide the full within-algorithm validator reliability analyses, reported per algorithm and per scenario, in Section~\ref{sec:validator_reliability}.

\subsection{Full Results on the Complete UDA Pipeline}
\label{sec:full_results}
This section provides the full per-scenario complete UDA results. Each table reports,
for one cross-domain scenario, the target accuracy of the checkpoint selected by every
validator (rows) for every algorithm (columns), along with \textit{Oracle},
\textit{SourceOnly}, and \textit{TargetOnly} references. Each table reports mean$\pm$std.
Tables~\ref{tab:adni-1-2}--\ref{tab:slo-oct} cover the eleven clinically relevant
transfer scenarios. Brain MRI: ADNI-1$\rightarrow$ADNI-2 (Table~\ref{tab:adni-1-2}),
ADNI-1$\rightarrow$ADNI-3 (Table~\ref{tab:adni-1-3}), ADNI-2$\rightarrow$ADNI-1
(Table~\ref{tab:adni-2-1}), ADNI-2$\rightarrow$ADNI-3 (Table~\ref{tab:adni-2-3}),
ADNI-1+2$\rightarrow$AIBL (Table~\ref{tab:adni-aibl}). Chest X-ray:
RSNA$\rightarrow$Child CXR (Table~\ref{tab:rsna-pedia-resnet-s}), Child
CXR$\rightarrow$RSNA (Table~\ref{tab:pedia-rsna-resnet}), LDD$\rightarrow$CRD
(Table~\ref{tab:ldd-crd-resnet}), CRD$\rightarrow$LDD (Table~\ref{tab:crd-ldd-resnet}).
Retinal: OCT$\rightarrow$SLO (Table~\ref{tab:oct-slo}), SLO$\rightarrow$OCT
(Table~\ref{tab:slo-oct}).

\begin{table*}[h]
\centering
\caption{Complete UDA results on ADNI-1$\rightarrow$ADNI-2 (target accuracy, \%; mean with std). Each cell reports the mean target accuracy (std) of the checkpoint a validator selects for an algorithm. The per-algorithm columns report selection within a single algorithm, \textbf{Avg.} is their mean, and \textbf{Across-Algo} pools the checkpoints of all algorithms and selects across them. \textit{Oracle} selects using target labels and is the actual best model, \textit{SourceOnly} is trained on source data only, and \textit{TargetOnly} is trained on labeled target data.}
\label{tab:adni-1-2}
\footnotesize
\setlength{\tabcolsep}{3pt}
\renewcommand{\arraystretch}{1.0}
\resizebox{\textwidth}{!}{%
\begin{tabular}{l | c | *{9}{c} : c | c | c}
\hline
 & \textit{SourceOnly} & {MMD} & {DANN} & {CDAN} & {DALN} & {MCC} & {BNM} & {ATDOC} & {MCD} & {AD2A} & \textbf{Avg.} & \textbf{Across-Algo} & \textit{TargetOnly} \\
\hline
\textit{Oracle} & 88.2\tiny$\pm$2.0 & 88.0\tiny$\pm$3.3 & 90.0\tiny$\pm$4.0 & 89.9\tiny$\pm$3.0 & 90.5\tiny$\pm$6.8 & 89.9\tiny$\pm$1.3 & 89.8\tiny$\pm$4.0 & 89.5\tiny$\pm$1.5 & 87.4\tiny$\pm$4.3 & 92.1\tiny$\pm$0.93 & 89.7\tiny$\pm$1.3 & 93.2\tiny$\pm$2.0 & 90.9\tiny$\pm$2.4 \\
\hline
Source-Risk & 81.2\tiny$\pm$5.8 & 82.8\tiny$\pm$3.1 & 81.8\tiny$\pm$4.0 & 81.1\tiny$\pm$8.0 & 85.9\tiny$\pm$7.4 & 81.6\tiny$\pm$2.5 & 82.9\tiny$\pm$3.7 & 79.0\tiny$\pm$6.2 & 81.5\tiny$\pm$6.3 & 86.8\tiny$\pm$2.8 & - & 84.4\tiny$\pm$2.5 & - \\
IWCV & 76.8\tiny$\pm$9.7 & 78.7\tiny$\pm$5.9 & 74.6\tiny$\pm$12 & 81.1\tiny$\pm$6.4 & 82.3\tiny$\pm$7.3 & 72.2\tiny$\pm$12 & 76.8\tiny$\pm$4.3 & 76.2\tiny$\pm$2.3 & 80.1\tiny$\pm$11 & 84.3\tiny$\pm$6.4 & - & 76.7\tiny$\pm$4.4 & - \\
DEV & 78.0\tiny$\pm$11 & 81.5\tiny$\pm$3.2 & 84.5\tiny$\pm$6.2 & 82.0\tiny$\pm$6.7 & 84.0\tiny$\pm$10 & 70.8\tiny$\pm$9.6 & 74.0\tiny$\pm$6.9 & 71.2\tiny$\pm$21 & 68.6\tiny$\pm$19 & 81.4\tiny$\pm$8.2 & - & 84.4\tiny$\pm$5.9 & - \\
DEV-N & 80.9\tiny$\pm$5.8 & 82.4\tiny$\pm$2.5 & 84.8\tiny$\pm$2.0 & 81.8\tiny$\pm$8.4 & 84.9\tiny$\pm$8.7 & 81.8\tiny$\pm$2.8 & 81.5\tiny$\pm$0.82 & 79.0\tiny$\pm$6.2 & 81.0\tiny$\pm$6.2 & 87.6\tiny$\pm$1.9 & - & 85.7\tiny$\pm$1.2 & - \\
Entropy & 60.9\tiny$\pm$15 & 49.9\tiny$\pm$0.30 & 56.7\tiny$\pm$15 & 57.0\tiny$\pm$16 & 50.0\tiny$\pm$0.0 & 61.1\tiny$\pm$19 & 59.4\tiny$\pm$18 & 50.0\tiny$\pm$0.0 & 70.1\tiny$\pm$19 & 56.9\tiny$\pm$15 & - & 50.0\tiny$\pm$0.0 & - \\
InfoMax & 84.2\tiny$\pm$2.1 & 82.4\tiny$\pm$2.1 & 87.3\tiny$\pm$4.4 & 86.0\tiny$\pm$1.5 & 89.5\tiny$\pm$6.7 & 76.2\tiny$\pm$8.1 & 84.6\tiny$\pm$7.4 & 80.2\tiny$\pm$11 & 83.3\tiny$\pm$5.6 & 86.3\tiny$\pm$2.2 & - & 83.2\tiny$\pm$7.2 & - \\
Corr-C & 84.6\tiny$\pm$4.1 & 81.9\tiny$\pm$3.1 & 83.4\tiny$\pm$3.6 & 83.1\tiny$\pm$3.6 & 84.8\tiny$\pm$9.3 & 75.3\tiny$\pm$5.8 & 80.7\tiny$\pm$3.9 & 84.7\tiny$\pm$2.6 & 84.1\tiny$\pm$4.4 & 85.6\tiny$\pm$5.1 & - & 79.6\tiny$\pm$3.0 & - \\
MCC (V) & 84.2\tiny$\pm$2.7 & 81.9\tiny$\pm$3.6 & 84.9\tiny$\pm$3.2 & 84.5\tiny$\pm$4.0 & 85.0\tiny$\pm$6.5 & 76.2\tiny$\pm$8.1 & 87.1\tiny$\pm$2.7 & 87.9\tiny$\pm$3.8 & 83.1\tiny$\pm$3.7 & 87.1\tiny$\pm$3.6 & - & 80.4\tiny$\pm$8.3 & - \\
BNM (V) & 84.7\tiny$\pm$2.1 & 82.4\tiny$\pm$2.1 & 87.5\tiny$\pm$4.0 & 84.9\tiny$\pm$3.8 & 87.2\tiny$\pm$3.4 & 76.2\tiny$\pm$8.1 & 84.6\tiny$\pm$7.4 & 80.2\tiny$\pm$11 & 83.3\tiny$\pm$5.6 & 86.3\tiny$\pm$2.2 & - & 81.8\tiny$\pm$9.6 & - \\
ClassAMI & 80.9\tiny$\pm$6.4 & 58.1\tiny$\pm$6.8 & 82.1\tiny$\pm$3.1 & 82.7\tiny$\pm$3.8 & 84.0\tiny$\pm$6.4 & 82.0\tiny$\pm$1.5 & 80.0\tiny$\pm$2.3 & 85.1\tiny$\pm$1.1 & 59.2\tiny$\pm$9.8 & 81.6\tiny$\pm$2.1 & - & 81.0\tiny$\pm$2.2 & - \\
SND & 73.9\tiny$\pm$12 & 84.2\tiny$\pm$4.2 & 80.0\tiny$\pm$2.6 & 79.2\tiny$\pm$5.8 & 81.8\tiny$\pm$3.9 & 77.6\tiny$\pm$2.3 & 77.3\tiny$\pm$9.4 & 83.6\tiny$\pm$2.7 & 82.2\tiny$\pm$5.6 & 80.6\tiny$\pm$5.3 & - & 80.7\tiny$\pm$8.7 & - \\
MixVal & 78.5\tiny$\pm$15 & 77.4\tiny$\pm$3.8 & 69.8\tiny$\pm$15 & 74.6\tiny$\pm$13 & 84.1\tiny$\pm$16 & 63.5\tiny$\pm$19 & 78.9\tiny$\pm$19 & 55.9\tiny$\pm$13 & 63.2\tiny$\pm$17 & 58.7\tiny$\pm$14 & - & 72.5\tiny$\pm$20 & - \\
TransScore & 85.2\tiny$\pm$2.8 & 82.7\tiny$\pm$5.5 & 86.9\tiny$\pm$2.2 & 85.9\tiny$\pm$2.0 & 68.1\tiny$\pm$20 & 78.1\tiny$\pm$9.2 & 76.3\tiny$\pm$11 & 77.3\tiny$\pm$6.5 & 83.3\tiny$\pm$5.4 & 87.8\tiny$\pm$3.1 & - & 78.5\tiny$\pm$11 & - \\

\hline
\end{tabular}%
}
\end{table*}

\begin{table*}[h]
\centering
\caption{Complete UDA results on ADNI-1$\rightarrow$ADNI-3 (target accuracy, \%; mean with std). Each cell reports the mean target accuracy (std) of the checkpoint a validator selects for an algorithm. The per-algorithm columns report selection within a single algorithm, \textbf{Avg.} is their mean, and \textbf{Across-Algo} pools the checkpoints of all algorithms and selects across them. \textit{Oracle} selects using target labels and is the actual best model, \textit{SourceOnly} is trained on source data only, and \textit{TargetOnly} is trained on labeled target data.}
\label{tab:adni-1-3}
\footnotesize
\setlength{\tabcolsep}{3pt}
\renewcommand{\arraystretch}{1.0}
\resizebox{\textwidth}{!}{%
\begin{tabular}{l | c | *{9}{c} : c | c | c}
\hline
 & \textit{SourceOnly} & {MMD} & {DANN} & {CDAN} & {DALN} & {MCC} & {BNM} & {ATDOC} & {MCD} & {AD2A} & \textbf{Avg.} & \textbf{Across-Algo} & \textit{TargetOnly} \\
\hline
\textit{Oracle} & 90.2\tiny$\pm$3.7 & 87.3\tiny$\pm$4.5 & 90.7\tiny$\pm$3.1 & 93.5\tiny$\pm$3.5 & 90.8\tiny$\pm$4.1 & 93.7\tiny$\pm$2.4 & 89.0\tiny$\pm$5.6 & 85.0\tiny$\pm$6.2 & 90.4\tiny$\pm$3.0 & 91.1\tiny$\pm$4.3 & 90.2\tiny$\pm$2.6 & 94.2\tiny$\pm$3.1 & 90.2\tiny$\pm$4.6 \\
\hline
Source-Risk & 82.8\tiny$\pm$8.3 & 78.0\tiny$\pm$9.9 & 83.8\tiny$\pm$5.3 & 82.7\tiny$\pm$8.8 & 82.4\tiny$\pm$7.2 & 82.6\tiny$\pm$3.2 & 78.0\tiny$\pm$7.1 & 75.9\tiny$\pm$14 & 86.3\tiny$\pm$1.6 & 83.0\tiny$\pm$6.7 & - & 83.5\tiny$\pm$4.8 & - \\
IWCV & 70.0\tiny$\pm$10 & 76.7\tiny$\pm$11 & 79.6\tiny$\pm$4.9 & 84.1\tiny$\pm$5.9 & 74.7\tiny$\pm$15 & 82.7\tiny$\pm$2.3 & 56.5\tiny$\pm$9.1 & 67.4\tiny$\pm$15 & 82.9\tiny$\pm$6.3 & 75.3\tiny$\pm$12 & - & 64.9\tiny$\pm$15 & - \\
DEV & 73.9\tiny$\pm$12 & 69.4\tiny$\pm$11 & 82.2\tiny$\pm$4.9 & 86.5\tiny$\pm$5.0 & 81.1\tiny$\pm$7.2 & 83.7\tiny$\pm$0.92 & 62.9\tiny$\pm$13 & 73.2\tiny$\pm$16 & 83.2\tiny$\pm$6.8 & 81.6\tiny$\pm$5.8 & - & 82.3\tiny$\pm$7.1 & - \\
DEV-N & 83.3\tiny$\pm$7.1 & 79.2\tiny$\pm$9.7 & 83.8\tiny$\pm$5.3 & 82.7\tiny$\pm$8.8 & 82.4\tiny$\pm$5.0 & 81.8\tiny$\pm$2.3 & 76.1\tiny$\pm$8.5 & 73.6\tiny$\pm$16 & 82.0\tiny$\pm$3.6 & 83.2\tiny$\pm$4.4 & - & 84.5\tiny$\pm$5.0 & - \\
Entropy & 67.8\tiny$\pm$13 & 66.8\tiny$\pm$15 & 57.2\tiny$\pm$15 & 57.3\tiny$\pm$14 & 63.2\tiny$\pm$18 & 69.2\tiny$\pm$27 & 64.0\tiny$\pm$11 & 53.1\tiny$\pm$13 & 61.8\tiny$\pm$19 & 50.3\tiny$\pm$0.72 & - & 52.9\tiny$\pm$6.5 & - \\
InfoMax & 79.0\tiny$\pm$5.9 & 72.0\tiny$\pm$14 & 79.7\tiny$\pm$5.0 & 78.2\tiny$\pm$4.1 & 79.4\tiny$\pm$6.8 & 75.5\tiny$\pm$4.3 & 72.7\tiny$\pm$6.5 & 68.8\tiny$\pm$15 & 83.6\tiny$\pm$4.5 & 80.2\tiny$\pm$5.7 & - & 70.6\tiny$\pm$7.9 & - \\
Corr-C & 79.7\tiny$\pm$6.7 & 70.8\tiny$\pm$5.1 & 76.8\tiny$\pm$1.5 & 75.9\tiny$\pm$3.7 & 74.5\tiny$\pm$4.3 & 74.2\tiny$\pm$2.5 & 69.9\tiny$\pm$5.0 & 60.9\tiny$\pm$13 & 77.7\tiny$\pm$2.8 & 72.1\tiny$\pm$11 & - & 68.7\tiny$\pm$8.5 & - \\
MCC (V) & 79.1\tiny$\pm$7.9 & 74.9\tiny$\pm$14 & 84.6\tiny$\pm$2.3 & 84.5\tiny$\pm$4.0 & 79.9\tiny$\pm$6.3 & 85.9\tiny$\pm$3.5 & 72.4\tiny$\pm$4.2 & 68.1\tiny$\pm$23 & 86.4\tiny$\pm$2.2 & 83.5\tiny$\pm$7.6 & - & 75.9\tiny$\pm$9.8 & - \\
BNM (V) & 78.2\tiny$\pm$7.3 & 75.0\tiny$\pm$15 & 79.7\tiny$\pm$5.0 & 78.2\tiny$\pm$4.1 & 79.4\tiny$\pm$6.8 & 75.5\tiny$\pm$4.3 & 72.7\tiny$\pm$6.5 & 71.9\tiny$\pm$18 & 83.6\tiny$\pm$4.6 & 80.2\tiny$\pm$5.7 & - & 70.6\tiny$\pm$7.9 & - \\
ClassAMI & 80.7\tiny$\pm$6.5 & 54.6\tiny$\pm$18 & 83.6\tiny$\pm$2.0 & 82.9\tiny$\pm$2.2 & 80.6\tiny$\pm$6.7 & 87.0\tiny$\pm$0.92 & 81.3\tiny$\pm$7.1 & 69.3\tiny$\pm$9.8 & 59.5\tiny$\pm$14 & 80.1\tiny$\pm$7.3 & - & 80.6\tiny$\pm$3.6 & - \\
SND & 73.8\tiny$\pm$6.0 & 71.6\tiny$\pm$5.0 & 77.5\tiny$\pm$4.5 & 75.4\tiny$\pm$6.0 & 76.8\tiny$\pm$4.3 & 70.8\tiny$\pm$9.4 & 71.2\tiny$\pm$9.0 & 66.7\tiny$\pm$11 & 76.7\tiny$\pm$3.8 & 74.5\tiny$\pm$2.0 & - & 76.2\tiny$\pm$3.9 & - \\
MixVal & 77.8\tiny$\pm$13 & 65.5\tiny$\pm$6.2 & 73.6\tiny$\pm$12 & 78.8\tiny$\pm$15 & 65.6\tiny$\pm$9.6 & 85.5\tiny$\pm$3.5 & 71.9\tiny$\pm$13 & 67.0\tiny$\pm$0.53 & 71.6\tiny$\pm$14 & 64.6\tiny$\pm$14 & - & 64.5\tiny$\pm$9.1 & - \\
TransScore & 76.3\tiny$\pm$6.4 & 69.9\tiny$\pm$13 & 81.7\tiny$\pm$2.7 & 81.1\tiny$\pm$5.9 & 76.9\tiny$\pm$5.0 & 74.2\tiny$\pm$2.5 & 47.3\tiny$\pm$19 & 71.2\tiny$\pm$17 & 75.9\tiny$\pm$6.4 & 72.9\tiny$\pm$22 & - & 62.6\tiny$\pm$16 & - \\
\hline
\end{tabular}%
}
\end{table*}

\begin{table*}[h]
\centering
\caption{Complete UDA results on ADNI-2$\rightarrow$ADNI-1 (target accuracy, \%; mean with std). Each cell reports the mean target accuracy (std) of the checkpoint a validator selects for an algorithm. The per-algorithm columns report selection within a single algorithm, \textbf{Avg.} is their mean, and \textbf{Across-Algo} pools the checkpoints of all algorithms and selects across them. \textit{Oracle} selects using target labels and is the actual best model, \textit{SourceOnly} is trained on source data only, and \textit{TargetOnly} is trained on labeled target data.}
\label{tab:adni-2-1}
\footnotesize
\setlength{\tabcolsep}{3pt}
\renewcommand{\arraystretch}{1.0}
\resizebox{\textwidth}{!}{%
\begin{tabular}{l | c | *{9}{c} : c | c | c}
\hline
 & \textit{SourceOnly} & {MMD} & {DANN} & {CDAN} & {DALN} & {MCC} & {BNM} & {ATDOC} & {MCD} & {AD2A} & \textbf{Avg.} & \textbf{Across-Algo} & \textit{TargetOnly} \\
\hline
\textit{Oracle} & 85.5\tiny$\pm$3.9 & 86.9\tiny$\pm$3.5 & 84.9\tiny$\pm$1.5 & 87.4\tiny$\pm$2.8 & 87.2\tiny$\pm$3.2 & 85.5\tiny$\pm$3.8 & 88.2\tiny$\pm$3.6 & 88.0\tiny$\pm$3.8 & 82.9\tiny$\pm$3.2 & 87.8\tiny$\pm$3.2 & 86.5\tiny$\pm$1.7 & 89.2\tiny$\pm$2.9 & 91.0\tiny$\pm$4.2 \\
\hline
Source-Risk & 77.4\tiny$\pm$5.0 & 81.0\tiny$\pm$5.2 & 75.2\tiny$\pm$2.2 & 75.2\tiny$\pm$7.5 & 79.8\tiny$\pm$5.4 & 80.2\tiny$\pm$6.3 & 77.5\tiny$\pm$5.7 & 81.4\tiny$\pm$5.1 & 74.4\tiny$\pm$3.2 & 81.5\tiny$\pm$5.1 & - & 78.2\tiny$\pm$4.3 & - \\
IWCV & 75.1\tiny$\pm$3.6 & 79.7\tiny$\pm$4.0 & 69.9\tiny$\pm$7.7 & 79.2\tiny$\pm$7.3 & 76.5\tiny$\pm$6.0 & 77.7\tiny$\pm$6.1 & 80.1\tiny$\pm$1.2 & 82.2\tiny$\pm$8.1 & 77.3\tiny$\pm$7.0 & 78.9\tiny$\pm$6.9 & - & 81.9\tiny$\pm$4.4 & - \\
DEV & 76.3\tiny$\pm$2.8 & 79.7\tiny$\pm$4.0 & 72.2\tiny$\pm$11 & 77.8\tiny$\pm$8.4 & 81.2\tiny$\pm$3.4 & 77.7\tiny$\pm$6.1 & 77.3\tiny$\pm$4.1 & 76.0\tiny$\pm$6.3 & 75.3\tiny$\pm$8.0 & 80.7\tiny$\pm$4.7 & - & 79.3\tiny$\pm$2.2 & - \\
DEV-N & 78.8\tiny$\pm$3.1 & 80.7\tiny$\pm$5.2 & 70.1\tiny$\pm$8.0 & 78.2\tiny$\pm$6.5 & 78.9\tiny$\pm$5.9 & 82.8\tiny$\pm$3.9 & 80.4\tiny$\pm$3.1 & 81.5\tiny$\pm$5.0 & 75.2\tiny$\pm$4.8 & 85.5\tiny$\pm$0.54 & - & 81.7\tiny$\pm$5.0 & - \\
Entropy & 53.9\tiny$\pm$6.1 & 58.2\tiny$\pm$16 & 53.3\tiny$\pm$4.7 & 64.2\tiny$\pm$19 & 50.5\tiny$\pm$1.2 & 67.5\tiny$\pm$18 & 70.4\tiny$\pm$18 & 59.9\tiny$\pm$20 & 58.3\tiny$\pm$11 & 50.0\tiny$\pm$0.0 & - & 50.0\tiny$\pm$0.0 & - \\
InfoMax & 81.0\tiny$\pm$4.3 & 79.2\tiny$\pm$8.0 & 80.7\tiny$\pm$3.8 & 80.3\tiny$\pm$4.5 & 81.5\tiny$\pm$2.2 & 82.0\tiny$\pm$4.4 & 80.1\tiny$\pm$8.4 & 82.3\tiny$\pm$6.3 & 75.9\tiny$\pm$5.4 & 84.8\tiny$\pm$6.2 & - & 82.0\tiny$\pm$4.4 & - \\
Corr-C & 79.6\tiny$\pm$5.5 & 80.8\tiny$\pm$5.0 & 81.3\tiny$\pm$4.6 & 80.0\tiny$\pm$5.1 & 81.5\tiny$\pm$3.5 & 80.8\tiny$\pm$5.9 & 77.6\tiny$\pm$6.2 & 82.1\tiny$\pm$5.5 & 77.7\tiny$\pm$3.0 & 86.1\tiny$\pm$2.0 & - & 80.8\tiny$\pm$5.9 & - \\
MCC (V) & 80.5\tiny$\pm$4.5 & 82.0\tiny$\pm$5.1 & 80.7\tiny$\pm$0.99 & 82.4\tiny$\pm$3.9 & 82.5\tiny$\pm$3.5 & 73.6\tiny$\pm$11 & 79.6\tiny$\pm$2.0 & 80.8\tiny$\pm$7.3 & 75.4\tiny$\pm$4.5 & 83.9\tiny$\pm$4.3 & - & 73.6\tiny$\pm$11 & - \\
BNM (V) & 80.9\tiny$\pm$4.1 & 79.5\tiny$\pm$7.8 & 80.7\tiny$\pm$3.8 & 80.7\tiny$\pm$4.4 & 82.0\tiny$\pm$3.1 & 82.0\tiny$\pm$4.4 & 80.1\tiny$\pm$8.4 & 82.3\tiny$\pm$6.3 & 75.2\tiny$\pm$4.8 & 84.4\tiny$\pm$6.2 & - & 82.0\tiny$\pm$4.4 & - \\
ClassAMI & 76.3\tiny$\pm$3.1 & 78.1\tiny$\pm$2.2 & 79.4\tiny$\pm$3.2 & 80.5\tiny$\pm$4.1 & 80.8\tiny$\pm$3.8 & 80.4\tiny$\pm$3.9 & 78.7\tiny$\pm$6.7 & 79.5\tiny$\pm$3.3 & 60.3\tiny$\pm$5.7 & 78.8\tiny$\pm$6.8 & - & 79.7\tiny$\pm$3.7 & - \\
SND & 81.9\tiny$\pm$7.2 & 77.9\tiny$\pm$9.2 & 74.9\tiny$\pm$1.8 & 77.4\tiny$\pm$6.1 & 78.8\tiny$\pm$4.6 & 67.5\tiny$\pm$7.8 & 79.7\tiny$\pm$8.2 & 78.9\tiny$\pm$5.3 & 74.3\tiny$\pm$3.4 & 72.5\tiny$\pm$1.5 & - & 74.6\tiny$\pm$8.2 & - \\
MixVal & 79.1\tiny$\pm$2.9 & 64.1\tiny$\pm$17 & 81.3\tiny$\pm$0.17 & 79.4\tiny$\pm$7.1 & 81.5\tiny$\pm$4.1 & 73.2\tiny$\pm$14 & 70.0\tiny$\pm$9.6 & 76.5\tiny$\pm$2.9 & 62.2\tiny$\pm$12 & 64.9\tiny$\pm$15 & - & 67.2\tiny$\pm$11 & - \\
TransScore & 78.3\tiny$\pm$5.6 & 79.6\tiny$\pm$4.0 & 79.0\tiny$\pm$1.4 & 80.5\tiny$\pm$4.7 & 81.8\tiny$\pm$3.3 & 77.2\tiny$\pm$3.5 & 79.3\tiny$\pm$9.0 & 82.6\tiny$\pm$6.2 & 72.2\tiny$\pm$12 & 82.7\tiny$\pm$3.6 & - & 78.4\tiny$\pm$5.0 & - \\
\hline
\end{tabular}%
}
\end{table*}

\begin{table*}[h]
\centering
\caption{Complete UDA results on ADNI-2$\rightarrow$ADNI-3 (target accuracy, \%; mean with std). Each cell reports the mean target accuracy (std) of the checkpoint a validator selects for an algorithm. The per-algorithm columns report selection within a single algorithm, \textbf{Avg.} is their mean, and \textbf{Across-Algo} pools the checkpoints of all algorithms and selects across them. \textit{Oracle} selects using target labels and is the actual best model, \textit{SourceOnly} is trained on source data only, and \textit{TargetOnly} is trained on labeled target data.}
\label{tab:adni-2-3}
\footnotesize
\setlength{\tabcolsep}{3pt}
\renewcommand{\arraystretch}{1.0}
\resizebox{\textwidth}{!}{%
\begin{tabular}{l | c | *{9}{c} : c | c | c}
\hline
 & \textit{SourceOnly} & {MMD} & {DANN} & {CDAN} & {DALN} & {MCC} & {BNM} & {ATDOC} & {MCD} & {AD2A} & \textbf{Avg.} & \textbf{Across-Algo} & \textit{TargetOnly} \\
\hline
\textit{Oracle} & 88.1\tiny$\pm$3.8 & 89.3\tiny$\pm$4.4 & 91.9\tiny$\pm$3.2 & 91.2\tiny$\pm$3.0 & 90.7\tiny$\pm$2.9 & 90.1\tiny$\pm$4.1 & 90.8\tiny$\pm$2.9 & 89.2\tiny$\pm$3.1 & 90.6\tiny$\pm$3.2 & 91.2\tiny$\pm$2.8 & 90.5\tiny$\pm$0.85 & 92.0\tiny$\pm$2.8 & 90.2\tiny$\pm$4.6 \\
\hline
Source-Risk & 82.3\tiny$\pm$7.3 & 82.4\tiny$\pm$5.8 & 81.5\tiny$\pm$7.6 & 84.9\tiny$\pm$5.9 & 82.5\tiny$\pm$2.8 & 82.6\tiny$\pm$6.4 & 79.1\tiny$\pm$8.5 & 81.3\tiny$\pm$5.4 & 84.0\tiny$\pm$6.8 & 85.2\tiny$\pm$6.4 & - & 81.7\tiny$\pm$8.8 & - \\
IWCV & 81.0\tiny$\pm$2.3 & 79.3\tiny$\pm$2.4 & 82.6\tiny$\pm$6.5 & 84.4\tiny$\pm$6.2 & 84.0\tiny$\pm$7.0 & 81.2\tiny$\pm$2.9 & 81.2\tiny$\pm$6.6 & 77.4\tiny$\pm$7.6 & 82.4\tiny$\pm$6.7 & 85.2\tiny$\pm$5.5 & - & 84.8\tiny$\pm$7.8 & - \\
DEV & 81.6\tiny$\pm$7.7 & 78.8\tiny$\pm$2.8 & 88.0\tiny$\pm$1.3 & 83.2\tiny$\pm$5.2 & 84.9\tiny$\pm$6.3 & 78.8\tiny$\pm$6.6 & 82.5\tiny$\pm$7.4 & 77.3\tiny$\pm$4.0 & 83.5\tiny$\pm$6.5 & 83.2\tiny$\pm$5.2 & - & 82.3\tiny$\pm$7.7 & - \\
DEV-N & 82.0\tiny$\pm$7.2 & 80.7\tiny$\pm$3.5 & 81.9\tiny$\pm$8.2 & 84.1\tiny$\pm$6.5 & 83.7\tiny$\pm$3.4 & 83.2\tiny$\pm$5.8 & 81.0\tiny$\pm$6.5 & 80.2\tiny$\pm$6.8 & 83.4\tiny$\pm$6.4 & 85.5\tiny$\pm$5.9 & - & 80.9\tiny$\pm$9.1 & - \\
Entropy & 57.0\tiny$\pm$16 & 50.0\tiny$\pm$0.0 & 69.8\tiny$\pm$23 & 58.4\tiny$\pm$19 & 49.9\tiny$\pm$0.26 & 50.0\tiny$\pm$0.0 & 67.7\tiny$\pm$21 & 69.0\tiny$\pm$18 & 66.8\tiny$\pm$18 & 51.9\tiny$\pm$4.3 & - & 50.0\tiny$\pm$0.0 & - \\
InfoMax & 81.5\tiny$\pm$4.8 & 82.3\tiny$\pm$5.8 & 82.9\tiny$\pm$5.4 & 82.7\tiny$\pm$4.7 & 78.9\tiny$\pm$6.3 & 81.7\tiny$\pm$3.8 & 80.3\tiny$\pm$3.5 & 82.6\tiny$\pm$7.0 & 80.3\tiny$\pm$8.9 & 84.8\tiny$\pm$6.5 & - & 81.4\tiny$\pm$3.1 & - \\
Corr-C & 74.8\tiny$\pm$10 & 77.1\tiny$\pm$3.1 & 73.6\tiny$\pm$5.3 & 76.7\tiny$\pm$1.5 & 72.3\tiny$\pm$6.6 & 76.0\tiny$\pm$4.9 & 77.6\tiny$\pm$2.6 & 72.4\tiny$\pm$2.6 & 77.8\tiny$\pm$7.3 & 79.7\tiny$\pm$5.4 & - & 76.4\tiny$\pm$2.8 & - \\
MCC (V) & 83.8\tiny$\pm$6.7 & 81.7\tiny$\pm$5.6 & 88.8\tiny$\pm$5.1 & 85.6\tiny$\pm$5.8 & 84.6\tiny$\pm$6.8 & 84.7\tiny$\pm$6.3 & 82.5\tiny$\pm$7.3 & 78.6\tiny$\pm$7.2 & 82.0\tiny$\pm$6.9 & 74.4\tiny$\pm$19 & - & 81.8\tiny$\pm$3.9 & - \\
BNM (V) & 81.2\tiny$\pm$5.2 & 82.3\tiny$\pm$5.8 & 85.2\tiny$\pm$1.4 & 82.7\tiny$\pm$4.7 & 78.9\tiny$\pm$6.3 & 81.7\tiny$\pm$3.8 & 80.3\tiny$\pm$3.5 & 54.4\tiny$\pm$7.5 & 55.9\tiny$\pm$15 & 83.9\tiny$\pm$6.0 & - & 54.5\tiny$\pm$8.7 & - \\
ClassAMI & 81.6\tiny$\pm$6.9 & 80.5\tiny$\pm$3.3 & 83.5\tiny$\pm$4.9 & 82.4\tiny$\pm$4.4 & 83.4\tiny$\pm$7.4 & 81.5\tiny$\pm$6.3 & 83.7\tiny$\pm$5.5 & 79.1\tiny$\pm$6.1 & 69.1\tiny$\pm$11 & 84.8\tiny$\pm$4.3 & - & 81.6\tiny$\pm$6.3 & - \\
SND & 76.0\tiny$\pm$5.1 & 76.7\tiny$\pm$2.1 & 73.4\tiny$\pm$4.6 & 71.7\tiny$\pm$3.9 & 74.9\tiny$\pm$8.2 & 74.8\tiny$\pm$2.0 & 75.0\tiny$\pm$2.4 & 76.5\tiny$\pm$4.5 & 78.0\tiny$\pm$6.2 & 75.3\tiny$\pm$9.9 & - & 70.8\tiny$\pm$6.5 & - \\
MixVal & 76.5\tiny$\pm$15 & 68.0\tiny$\pm$11 & 74.3\tiny$\pm$14 & 74.4\tiny$\pm$12 & 59.9\tiny$\pm$9.2 & 80.2\tiny$\pm$9.0 & 65.1\tiny$\pm$18 & 67.7\tiny$\pm$16 & 75.0\tiny$\pm$13 & 58.8\tiny$\pm$9.0 & - & 52.5\tiny$\pm$3.4 & - \\
TransScore & 82.0\tiny$\pm$4.4 & 83.5\tiny$\pm$7.0 & 82.9\tiny$\pm$5.4 & 80.8\tiny$\pm$4.8 & 83.2\tiny$\pm$8.2 & 80.9\tiny$\pm$5.0 & 78.0\tiny$\pm$7.0 & 74.1\tiny$\pm$7.0 & 76.5\tiny$\pm$5.5 & 78.4\tiny$\pm$18 & - & 81.4\tiny$\pm$3.1 & - \\
\hline
\end{tabular}%
}
\end{table*}

\begin{table*}[h]
\centering
\caption{Complete UDA results on ADNI-1+2$\rightarrow$AIBL (target accuracy, \%; mean with std). Each cell reports the mean target accuracy (std) of the checkpoint a validator selects for an algorithm. The per-algorithm columns report selection within a single algorithm, \textbf{Avg.} is their mean, and \textbf{Across-Algo} pools the checkpoints of all algorithms and selects across them. \textit{Oracle} selects using target labels and is the actual best model, \textit{SourceOnly} is trained on source data only, and \textit{TargetOnly} is trained on labeled target data.}
\label{tab:adni-aibl}
\footnotesize
\setlength{\tabcolsep}{3pt}
\renewcommand{\arraystretch}{1.0}
\resizebox{\textwidth}{!}{%
\begin{tabular}{l | c | *{9}{c} : c | c | c}
\hline
 & \textit{SourceOnly} & {MMD} & {DANN} & {CDAN} & {DALN} & {MCC} & {BNM} & {ATDOC} & {MCD} & {AD2A} & \textbf{Avg.} & \textbf{Across-Algo} & \textit{TargetOnly} \\
\hline
\textit{Oracle} & 92.2\tiny$\pm$2.3 & 91.6\tiny$\pm$2.0 & 92.8\tiny$\pm$1.1 & 93.4\tiny$\pm$1.7 & 93.5\tiny$\pm$2.0 & 91.8\tiny$\pm$1.6 & 91.4\tiny$\pm$1.9 & 90.6\tiny$\pm$2.2 & 92.0\tiny$\pm$2.0 & 94.1\tiny$\pm$3.7 & 92.4\tiny$\pm$1.1 & 93.8\tiny$\pm$2.1 & 81.9\tiny$\pm$3.3 \\
\hline
Source-Risk & 88.4\tiny$\pm$3.0 & 85.0\tiny$\pm$3.5 & 86.7\tiny$\pm$4.5 & 87.8\tiny$\pm$3.9 & 86.7\tiny$\pm$9.1 & 87.6\tiny$\pm$2.7 & 84.0\tiny$\pm$3.2 & 81.2\tiny$\pm$4.7 & 84.9\tiny$\pm$4.2 & 88.5\tiny$\pm$0.95 & - & 84.0\tiny$\pm$5.1 & - \\
IWCV & 86.8\tiny$\pm$3.3 & 83.2\tiny$\pm$4.6 & 83.8\tiny$\pm$4.7 & 84.3\tiny$\pm$6.0 & 84.0\tiny$\pm$9.2 & 86.4\tiny$\pm$4.5 & 83.3\tiny$\pm$3.2 & 84.1\tiny$\pm$1.3 & 81.7\tiny$\pm$2.4 & 87.3\tiny$\pm$3.4 & - & 84.4\tiny$\pm$3.3 & - \\
DEV & 87.3\tiny$\pm$2.7 & 84.2\tiny$\pm$5.8 & 83.4\tiny$\pm$5.2 & 82.2\tiny$\pm$12 & 89.3\tiny$\pm$2.0 & 85.4\tiny$\pm$4.8 & 87.4\tiny$\pm$2.2 & 84.3\tiny$\pm$1.9 & 83.2\tiny$\pm$4.3 & 83.3\tiny$\pm$2.3 & - & 89.0\tiny$\pm$1.7 & - \\
DEV-N & 87.6\tiny$\pm$4.2 & 85.0\tiny$\pm$3.5 & 85.0\tiny$\pm$5.0 & 88.9\tiny$\pm$3.5 & 86.7\tiny$\pm$9.1 & 88.2\tiny$\pm$2.3 & 84.0\tiny$\pm$3.2 & 81.2\tiny$\pm$4.7 & 83.4\tiny$\pm$4.7 & 91.4\tiny$\pm$5.3 & - & 82.7\tiny$\pm$5.0 & - \\
Entropy & 88.3\tiny$\pm$3.3 & 84.5\tiny$\pm$5.4 & 88.6\tiny$\pm$4.8 & 86.9\tiny$\pm$8.0 & 61.7\tiny$\pm$21 & 66.7\tiny$\pm$15 & 79.0\tiny$\pm$16 & 81.5\tiny$\pm$8.8 & 77.1\tiny$\pm$18 & 72.0\tiny$\pm$31 & - & 64.8\tiny$\pm$16 & - \\
InfoMax & 84.7\tiny$\pm$6.7 & 82.4\tiny$\pm$4.9 & 78.9\tiny$\pm$3.2 & 84.2\tiny$\pm$5.0 & 82.2\tiny$\pm$4.2 & 79.4\tiny$\pm$5.5 & 81.6\tiny$\pm$5.0 & 84.0\tiny$\pm$5.7 & 85.5\tiny$\pm$4.8 & 85.6\tiny$\pm$1.8 & - & 81.6\tiny$\pm$5.0 & - \\
Corr-C & 80.4\tiny$\pm$6.8 & 77.3\tiny$\pm$4.7 & 75.4\tiny$\pm$3.0 & 77.5\tiny$\pm$3.6 & 80.0\tiny$\pm$3.1 & 74.3\tiny$\pm$4.7 & 75.0\tiny$\pm$3.4 & 77.2\tiny$\pm$3.6 & 77.0\tiny$\pm$4.1 & 76.2\tiny$\pm$2.3 & - & 75.1\tiny$\pm$3.2 & - \\
MCC (V) & 88.3\tiny$\pm$3.8 & 86.0\tiny$\pm$3.0 & 84.7\tiny$\pm$5.5 & 89.7\tiny$\pm$3.4 & 88.2\tiny$\pm$2.5 & 80.6\tiny$\pm$11 & 83.8\tiny$\pm$2.9 & 82.2\tiny$\pm$6.8 & 84.6\tiny$\pm$3.6 & 91.1\tiny$\pm$4.2 & - & 84.1\tiny$\pm$4.8 & - \\
BNM (V) & 88.2\tiny$\pm$4.1 & 84.7\tiny$\pm$3.0 & 80.2\tiny$\pm$1.1 & 83.9\tiny$\pm$5.2 & 85.6\tiny$\pm$4.0 & 79.4\tiny$\pm$5.5 & 81.6\tiny$\pm$5.0 & 62.7\tiny$\pm$5.3 & 73.3\tiny$\pm$14 & 90.2\tiny$\pm$4.7 & - & 65.7\tiny$\pm$8.2 & - \\
ClassAMI & 84.9\tiny$\pm$7.1 & 68.7\tiny$\pm$19 & 83.9\tiny$\pm$5.5 & 84.0\tiny$\pm$5.1 & 87.6\tiny$\pm$5.0 & 85.9\tiny$\pm$2.4 & 84.5\tiny$\pm$2.3 & 84.8\tiny$\pm$3.7 & 73.3\tiny$\pm$15 & 86.8\tiny$\pm$3.3 & - & 83.2\tiny$\pm$3.6 & - \\
SND & 74.0\tiny$\pm$1.8 & 75.2\tiny$\pm$3.8 & 76.0\tiny$\pm$5.6 & 77.9\tiny$\pm$3.6 & 77.6\tiny$\pm$7.5 & 75.2\tiny$\pm$4.7 & 74.1\tiny$\pm$5.7 & 74.8\tiny$\pm$2.3 & 78.2\tiny$\pm$5.1 & 76.8\tiny$\pm$5.1 & - & 77.2\tiny$\pm$5.1 & - \\
MixVal & 79.7\tiny$\pm$11 & 58.6\tiny$\pm$16 & 86.4\tiny$\pm$4.6 & 77.0\tiny$\pm$11 & 78.7\tiny$\pm$9.5 & 75.2\tiny$\pm$16 & 82.2\tiny$\pm$2.8 & 80.2\tiny$\pm$6.8 & 72.8\tiny$\pm$18 & 90.0\tiny$\pm$4.1 & - & 67.8\tiny$\pm$15 & - \\
TransScore & 88.5\tiny$\pm$4.0 & 82.6\tiny$\pm$5.8 & 79.5\tiny$\pm$3.9 & 82.1\tiny$\pm$6.7 & 83.0\tiny$\pm$5.8 & 80.7\tiny$\pm$6.0 & 85.0\tiny$\pm$1.3 & 81.2\tiny$\pm$6.6 & 81.2\tiny$\pm$4.1 & 85.6\tiny$\pm$1.8 & - & 80.1\tiny$\pm$4.7 & - \\
\hline
\end{tabular}%
}
\end{table*}

\begin{table*}[h]
\centering
\caption{Complete UDA results on RSNA$\rightarrow$Child CXR (target accuracy, \%; mean with std). Each cell reports the mean target accuracy (std) of the checkpoint a validator selects for an algorithm. The per-algorithm columns report selection within a single algorithm, \textbf{Avg.} is their mean, and \textbf{Across-Algo} pools the checkpoints of all algorithms and selects across them. \textit{Oracle} selects using target labels and is the actual best model, \textit{SourceOnly} is trained on source data only, and \textit{TargetOnly} is trained on labeled target data.}
\label{tab:rsna-pedia-resnet-s}
\footnotesize
\setlength{\tabcolsep}{3pt}
\renewcommand{\arraystretch}{1.0}
\resizebox{\textwidth}{!}{%
\begin{tabular}{l | c | *{9}{c} : c | c | c}
\hline
 & \textit{SourceOnly} & {MMD} & {DANN} & {CDAN} & {DALN} & {MCC} & {BNM} & {ATDOC} & {MCD} & {CoUDA} & \textbf{Avg.} & \textbf{Across-Algo} & \textit{TargetOnly} \\
\hline
\textit{Oracle} & 81.8\tiny$\pm$1.8 & 83.4\tiny$\pm$1.9 & 81.3\tiny$\pm$0.24 & 85.0\tiny$\pm$2.2 & 82.9\tiny$\pm$2.7 & 87.0\tiny$\pm$1.5 & 83.8\tiny$\pm$0.36 & 81.7\tiny$\pm$2.6 & 84.2\tiny$\pm$1.6 & 89.8\tiny$\pm$0.47 & 84.3\tiny$\pm$2.5 & 89.7\tiny$\pm$0.40 & 92.7\tiny$\pm$1.1 \\
\hline
Source-Risk & 74.0\tiny$\pm$2.6 & 72.2\tiny$\pm$1.4 & 74.4\tiny$\pm$0.06 & 74.2\tiny$\pm$4.7 & 72.7\tiny$\pm$5.2 & 83.1\tiny$\pm$2.7 & 70.4\tiny$\pm$4.9 & 46.5\tiny$\pm$16 & 76.6\tiny$\pm$4.2 & 84.9\tiny$\pm$4.3 & - & 75.3\tiny$\pm$4.1 & - \\
IWCV & 75.6\tiny$\pm$2.3 & 72.5\tiny$\pm$5.3 & 73.7\tiny$\pm$0.42 & 72.8\tiny$\pm$3.5 & 72.3\tiny$\pm$3.7 & 75.6\tiny$\pm$7.0 & 73.5\tiny$\pm$0.18 & 61.5\tiny$\pm$5.7 & 78.8\tiny$\pm$2.6 & 87.0\tiny$\pm$1.3 & - & 76.4\tiny$\pm$6.3 & - \\
DEV & 74.5\tiny$\pm$4.3 & 75.9\tiny$\pm$7.8 & 71.3\tiny$\pm$4.0 & 72.4\tiny$\pm$2.9 & 65.1\tiny$\pm$4.2 & 65.5\tiny$\pm$10 & 69.8\tiny$\pm$2.0 & 54.6\tiny$\pm$21 & 66.3\tiny$\pm$11 & 76.5\tiny$\pm$4.7 & - & 72.6\tiny$\pm$12 & - \\
DEV-N & 71.9\tiny$\pm$8.8 & 71.2\tiny$\pm$4.0 & 73.5\tiny$\pm$0.51 & 70.3\tiny$\pm$0.65 & 73.2\tiny$\pm$8.9 & 82.8\tiny$\pm$2.7 & 73.2\tiny$\pm$2.4 & 64.2\tiny$\pm$13 & 73.6\tiny$\pm$2.1 & 83.5\tiny$\pm$3.9 & - & 76.4\tiny$\pm$4.1 & - \\
Entropy & 64.9\tiny$\pm$17 & 59.7\tiny$\pm$10 & 59.2\tiny$\pm$3.6 & 59.7\tiny$\pm$8.4 & 53.9\tiny$\pm$5.5 & 57.3\tiny$\pm$13 & 51.3\tiny$\pm$0.27 & 62.1\tiny$\pm$5.6 & 66.9\tiny$\pm$15 & 72.0\tiny$\pm$19 & - & 51.0\tiny$\pm$0.68 & - \\
InfoMax & 72.5\tiny$\pm$1.6 & 73.2\tiny$\pm$2.3 & 77.2\tiny$\pm$2.0 & 79.0\tiny$\pm$3.3 & 72.0\tiny$\pm$5.7 & 82.0\tiny$\pm$0.74 & 71.5\tiny$\pm$4.3 & 52.7\tiny$\pm$15 & 76.1\tiny$\pm$0.57 & 87.8\tiny$\pm$1.6 & - & 78.8\tiny$\pm$5.9 & - \\
Corr-C & 72.5\tiny$\pm$1.6 & 75.3\tiny$\pm$5.6 & 72.2\tiny$\pm$9.9 & 78.7\tiny$\pm$2.7 & 72.0\tiny$\pm$2.7 & 82.2\tiny$\pm$0.56 & 70.6\tiny$\pm$3.1 & 55.0\tiny$\pm$16 & 64.3\tiny$\pm$28 & 87.9\tiny$\pm$1.4 & - & 79.1\tiny$\pm$6.2 & - \\
MCC (V) & 74.0\tiny$\pm$3.7 & 71.6\tiny$\pm$0.99 & 74.7\tiny$\pm$0.57 & 75.4\tiny$\pm$3.6 & 70.9\tiny$\pm$8.7 & 72.0\tiny$\pm$15 & 70.0\tiny$\pm$3.6 & 52.5\tiny$\pm$14 & 75.5\tiny$\pm$0.47 & 84.8\tiny$\pm$4.6 & - & 73.8\tiny$\pm$14 & - \\
BNM (V) & 72.5\tiny$\pm$1.6 & 73.2\tiny$\pm$2.3 & 77.2\tiny$\pm$2.0 & 71.2\tiny$\pm$17 & 72.0\tiny$\pm$5.7 & 82.5\tiny$\pm$1.2 & 71.5\tiny$\pm$4.3 & 52.7\tiny$\pm$15 & 76.1\tiny$\pm$0.59 & 87.5\tiny$\pm$1.4 & - & 78.8\tiny$\pm$5.9 & - \\
ClassAMI & 55.9\tiny$\pm$4.0 & 68.0\tiny$\pm$10 & 76.3\tiny$\pm$2.0 & 78.2\tiny$\pm$1.0 & 52.7\tiny$\pm$1.2 & 82.5\tiny$\pm$3.5 & 72.2\tiny$\pm$6.7 & 64.4\tiny$\pm$1.0 & 76.3\tiny$\pm$2.3 & 79.0\tiny$\pm$8.7 & - & 74.2\tiny$\pm$8.7 & - \\
SND & 70.5\tiny$\pm$1.4 & 74.8\tiny$\pm$4.6 & 69.2\tiny$\pm$5.8 & 75.7\tiny$\pm$6.9 & 66.2\tiny$\pm$1.1 & 71.0\tiny$\pm$3.1 & 70.8\tiny$\pm$2.0 & 58.0\tiny$\pm$12 & 80.2\tiny$\pm$2.1 & 85.5\tiny$\pm$1.7 & - & 70.5\tiny$\pm$9.7 & - \\
MixVal & 79.4\tiny$\pm$2.0 & 73.6\tiny$\pm$9.9 & 76.9\tiny$\pm$3.4 & 78.9\tiny$\pm$3.9 & 63.9\tiny$\pm$17 & 66.7\tiny$\pm$18 & 75.1\tiny$\pm$5.1 & 52.4\tiny$\pm$14 & 75.4\tiny$\pm$1.1 & 82.8\tiny$\pm$0.86 & - & 51.5\tiny$\pm$9.8 & - \\
TransScore & 73.7\tiny$\pm$0.51 & 75.2\tiny$\pm$2.9 & 75.4\tiny$\pm$0.82 & 76.1\tiny$\pm$1.8 & 70.9\tiny$\pm$2.9 & 69.4\tiny$\pm$13 & 69.8\tiny$\pm$1.9 & 53.1\tiny$\pm$14 & 76.1\tiny$\pm$0.59 & 87.8\tiny$\pm$0.17 & - & 70.3\tiny$\pm$12 & - \\
\hline
\end{tabular}%
}
\end{table*}

\begin{table*}[h]
\centering
\caption{Complete UDA results on Child CXR$\rightarrow$RSNA (target accuracy, \%; mean with std). Each cell reports the mean target accuracy (std) of the checkpoint a validator selects for an algorithm. The per-algorithm columns report selection within a single algorithm, \textbf{Avg.} is their mean, and \textbf{Across-Algo} pools the checkpoints of all algorithms and selects across them. \textit{Oracle} selects using target labels and is the actual best model, \textit{SourceOnly} is trained on source data only, and \textit{TargetOnly} is trained on labeled target data.}
\label{tab:pedia-rsna-resnet}
\footnotesize
\setlength{\tabcolsep}{3pt}
\renewcommand{\arraystretch}{1.0}
\resizebox{\textwidth}{!}{%
\begin{tabular}{l | c | *{9}{c} : c | c | c}
\hline
 & \textit{SourceOnly} & {MMD} & {DANN} & {CDAN} & {DALN} & {MCC} & {BNM} & {ATDOC} & {MCD} & {CoUDA} & \textbf{Avg.} & \textbf{Across-Algo} & \textit{TargetOnly} \\
\hline
\textit{Oracle} & 75.1\tiny$\pm$0.36 & 74.2\tiny$\pm$1.1 & 77.0\tiny$\pm$1.1 & 76.3\tiny$\pm$0.33 & 78.1\tiny$\pm$0.45 & 77.3\tiny$\pm$0.76 & 77.2\tiny$\pm$0.47 & 76.7\tiny$\pm$0.15 & 76.9\tiny$\pm$0.39 & 73.2\tiny$\pm$0.46 & 76.3\tiny$\pm$1.5 & 78.4\tiny$\pm$0.18 & 82.4\tiny$\pm$0.46 \\
\hline
Source-Risk & 72.4\tiny$\pm$2.8 & 67.0\tiny$\pm$5.0 & 68.5\tiny$\pm$3.0 & 71.7\tiny$\pm$1.3 & 74.6\tiny$\pm$2.1 & 73.5\tiny$\pm$2.1 & 71.8\tiny$\pm$3.1 & 46.7\tiny$\pm$20 & 67.1\tiny$\pm$4.9 & 71.2\tiny$\pm$0.98 & - & 72.2\tiny$\pm$1.2 & - \\
IWCV & 73.5\tiny$\pm$1.5 & 72.7\tiny$\pm$3.1 & 70.7\tiny$\pm$3.5 & 69.8\tiny$\pm$0.91 & 76.1\tiny$\pm$1.4 & 68.2\tiny$\pm$7.3 & 69.3\tiny$\pm$4.5 & 67.5\tiny$\pm$5.0 & 73.9\tiny$\pm$0.32 & 72.0\tiny$\pm$0.78 & - & 58.2\tiny$\pm$13 & - \\
DEV & 73.5\tiny$\pm$1.5 & 66.2\tiny$\pm$8.1 & 70.1\tiny$\pm$3.7 & 67.3\tiny$\pm$4.8 & 75.7\tiny$\pm$1.0 & 74.6\tiny$\pm$2.6 & 67.7\tiny$\pm$9.6 & 58.2\tiny$\pm$8.2 & 70.7\tiny$\pm$4.1 & 72.0\tiny$\pm$0.78 & - & 69.9\tiny$\pm$3.4 & - \\
DEV-N & 71.7\tiny$\pm$2.8 & 67.0\tiny$\pm$5.0 & 68.5\tiny$\pm$3.0 & 71.7\tiny$\pm$1.3 & 74.9\tiny$\pm$2.1 & 73.5\tiny$\pm$2.1 & 71.8\tiny$\pm$3.1 & 53.8\tiny$\pm$30 & 67.1\tiny$\pm$4.9 & 71.2\tiny$\pm$0.98 & - & 71.9\tiny$\pm$2.7 & - \\
Entropy & 72.4\tiny$\pm$1.1 & 67.5\tiny$\pm$1.7 & 68.8\tiny$\pm$1.7 & 65.8\tiny$\pm$4.1 & 76.2\tiny$\pm$1.6 & 73.8\tiny$\pm$1.4 & 72.8\tiny$\pm$3.4 & 62.3\tiny$\pm$8.6 & 75.4\tiny$\pm$0.94 & 71.2\tiny$\pm$0.42 & - & 74.3\tiny$\pm$1.8 & - \\
InfoMax & 72.5\tiny$\pm$1.0 & 70.1\tiny$\pm$0.97 & 73.4\tiny$\pm$1.1 & 74.4\tiny$\pm$1.1 & 75.7\tiny$\pm$1.2 & 73.8\tiny$\pm$2.0 & 74.1\tiny$\pm$0.98 & 54.0\tiny$\pm$22 & 73.8\tiny$\pm$0.38 & 71.2\tiny$\pm$0.42 & - & 74.3\tiny$\pm$2.3 & - \\
Corr-C & 72.3\tiny$\pm$0.96 & 73.3\tiny$\pm$2.3 & 76.1\tiny$\pm$1.1 & 75.9\tiny$\pm$0.38 & 76.0\tiny$\pm$0.51 & 75.3\tiny$\pm$1.2 & 73.9\tiny$\pm$0.92 & 69.4\tiny$\pm$3.3 & 73.1\tiny$\pm$0.39 & 71.5\tiny$\pm$0.57 & - & 74.4\tiny$\pm$1.6 & - \\
MCC (V) & 72.6\tiny$\pm$1.2 & 68.4\tiny$\pm$2.4 & 69.4\tiny$\pm$1.5 & 68.9\tiny$\pm$3.8 & 74.3\tiny$\pm$4.3 & 74.2\tiny$\pm$2.0 & 74.4\tiny$\pm$1.4 & 69.9\tiny$\pm$2.2 & 75.1\tiny$\pm$1.4 & 71.2\tiny$\pm$0.36 & - & 74.9\tiny$\pm$2.4 & - \\
BNM (V) & 72.5\tiny$\pm$1.0 & 70.1\tiny$\pm$0.97 & 73.4\tiny$\pm$1.1 & 74.4\tiny$\pm$1.1 & 75.7\tiny$\pm$1.2 & 73.8\tiny$\pm$2.0 & 74.0\tiny$\pm$0.98 & 54.0\tiny$\pm$22 & 73.8\tiny$\pm$0.38 & 71.2\tiny$\pm$0.42 & - & 74.3\tiny$\pm$2.3 & - \\
ClassAMI & 71.4\tiny$\pm$1.2 & 69.2\tiny$\pm$2.5 & 69.3\tiny$\pm$1.9 & 71.0\tiny$\pm$2.9 & 75.3\tiny$\pm$0.84 & 74.0\tiny$\pm$1.6 & 72.9\tiny$\pm$1.5 & 69.0\tiny$\pm$3.9 & 70.3\tiny$\pm$6.2 & 71.2\tiny$\pm$0.43 & - & 73.4\tiny$\pm$2.2 & - \\
SND & 73.0\tiny$\pm$1.4 & 74.1\tiny$\pm$1.1 & 73.8\tiny$\pm$6.1 & 76.3\tiny$\pm$0.33 & 74.7\tiny$\pm$1.2 & 72.7\tiny$\pm$2.2 & 71.6\tiny$\pm$4.3 & 59.9\tiny$\pm$24 & 72.4\tiny$\pm$2.4 & 72.1\tiny$\pm$0.33 & - & 74.6\tiny$\pm$1.4 & - \\
MixVal & 73.3\tiny$\pm$0.51 & 69.0\tiny$\pm$1.8 & 69.9\tiny$\pm$2.7 & 71.6\tiny$\pm$3.1 & 76.7\tiny$\pm$0.72 & 73.5\tiny$\pm$0.86 & 73.8\tiny$\pm$1.4 & 50.6\tiny$\pm$24 & 73.3\tiny$\pm$1.8 & 71.3\tiny$\pm$0.37 & - & 73.9\tiny$\pm$1.6 & - \\
TransScore & 70.5\tiny$\pm$2.8 & 69.8\tiny$\pm$0.83 & 70.6\tiny$\pm$1.5 & 71.9\tiny$\pm$2.4 & 76.8\tiny$\pm$0.93 & 73.4\tiny$\pm$0.98 & 73.5\tiny$\pm$2.2 & 62.3\tiny$\pm$9.8 & 74.6\tiny$\pm$0.65 & 71.3\tiny$\pm$0.37 & - & 75.4\tiny$\pm$1.0 & - \\
\hline
\end{tabular}%
}
\end{table*}

\begin{table*}[h]
\centering
\caption{Complete UDA results on LDD$\rightarrow$CRD (target accuracy, \%; mean with std). Each cell reports the mean target accuracy (std) of the checkpoint a validator selects for an algorithm. The per-algorithm columns report selection within a single algorithm, \textbf{Avg.} is their mean, and \textbf{Across-Algo} pools the checkpoints of all algorithms and selects across them. \textit{Oracle} selects using target labels and is the actual best model, \textit{SourceOnly} is trained on source data only, and \textit{TargetOnly} is trained on labeled target data.}
\label{tab:ldd-crd-resnet}
\footnotesize
\setlength{\tabcolsep}{3pt}
\renewcommand{\arraystretch}{1.0}
\resizebox{\textwidth}{!}{%
\begin{tabular}{l | c | *{9}{c} : c | c | c}
\hline
 & \textit{SourceOnly} & {MMD} & {DANN} & {CDAN} & {DALN} & {MCC} & {BNM} & {ATDOC} & {MCD} & {CoUDA} & \textbf{Avg.} & \textbf{Across-Algo} & \textit{TargetOnly} \\
\hline
\textit{Oracle} & 79.1\tiny$\pm$0.10 & 82.6\tiny$\pm$0.04 & 82.9\tiny$\pm$0.08 & 82.7\tiny$\pm$0.58 & 82.9\tiny$\pm$0.55 & 82.4\tiny$\pm$0.90 & 83.2\tiny$\pm$0.58 & 82.8\tiny$\pm$0.44 & 83.2\tiny$\pm$0.82 & 81.7\tiny$\pm$0.92 & 82.7\tiny$\pm$0.44 & 83.5\tiny$\pm$0.14 & 95.7\tiny$\pm$0.34 \\
\hline
Source-Risk & 66.8\tiny$\pm$2.9 & 80.5\tiny$\pm$0.46 & 79.4\tiny$\pm$3.8 & 69.9\tiny$\pm$6.2 & 79.7\tiny$\pm$2.7 & 74.1\tiny$\pm$2.6 & 81.0\tiny$\pm$2.2 & 77.8\tiny$\pm$3.8 & 79.4\tiny$\pm$1.3 & 74.8\tiny$\pm$1.6 & - & 79.3\tiny$\pm$0.68 & - \\
IWCV & 67.5\tiny$\pm$4.5 & 79.7\tiny$\pm$0.05 & 78.1\tiny$\pm$2.0 & 70.2\tiny$\pm$3.8 & 79.0\tiny$\pm$2.3 & 74.7\tiny$\pm$2.2 & 79.2\tiny$\pm$3.4 & 78.2\tiny$\pm$1.8 & 76.6\tiny$\pm$2.7 & 77.3\tiny$\pm$0.33 & - & 75.3\tiny$\pm$1.2 & - \\
DEV & 71.7\tiny$\pm$10 & 79.8\tiny$\pm$0.23 & 79.9\tiny$\pm$0.14 & 78.0\tiny$\pm$2.8 & 79.5\tiny$\pm$4.3 & 72.9\tiny$\pm$8.8 & 75.7\tiny$\pm$5.3 & 57.0\tiny$\pm$13 & 79.5\tiny$\pm$3.9 & 80.7\tiny$\pm$0.45 & - & 72.9\tiny$\pm$13 & - \\
DEV-N & 66.8\tiny$\pm$2.9 & 80.5\tiny$\pm$0.36 & 78.1\tiny$\pm$2.0 & 70.3\tiny$\pm$6.6 & 80.0\tiny$\pm$2.8 & 74.1\tiny$\pm$2.6 & 81.5\tiny$\pm$1.2 & 77.9\tiny$\pm$3.9 & 78.9\tiny$\pm$1.8 & 74.8\tiny$\pm$1.6 & - & 79.5\tiny$\pm$0.67 & - \\
Entropy & 64.2\tiny$\pm$0.85 & 80.0\tiny$\pm$0.24 & 81.3\tiny$\pm$0.44 & 73.0\tiny$\pm$2.4 & 73.2\tiny$\pm$16 & 70.5\tiny$\pm$12 & 58.0\tiny$\pm$16 & 71.3\tiny$\pm$15 & 80.5\tiny$\pm$1.9 & 75.2\tiny$\pm$0.16 & - & 50.0\tiny$\pm$0.01 & - \\
InfoMax & 77.6\tiny$\pm$1.1 & 81.2\tiny$\pm$1.9 & 81.7\tiny$\pm$1.2 & 80.8\tiny$\pm$0.69 & 82.0\tiny$\pm$0.93 & 81.5\tiny$\pm$1.4 & 81.8\tiny$\pm$0.52 & 80.3\tiny$\pm$1.5 & 81.8\tiny$\pm$1.1 & 79.1\tiny$\pm$1.1 & - & 81.8\tiny$\pm$0.63 & - \\
Corr-C & 78.0\tiny$\pm$1.6 & 79.9\tiny$\pm$0.50 & 82.1\tiny$\pm$0.83 & 79.5\tiny$\pm$1.9 & 82.0\tiny$\pm$0.92 & 79.2\tiny$\pm$2.6 & 80.7\tiny$\pm$1.6 & 78.0\tiny$\pm$6.9 & 81.8\tiny$\pm$1.9 & 78.5\tiny$\pm$2.2 & - & 80.4\tiny$\pm$1.8 & - \\
MCC (V) & 66.6\tiny$\pm$1.1 & 80.5\tiny$\pm$0.95 & 81.3\tiny$\pm$0.44 & 74.4\tiny$\pm$1.8 & 80.9\tiny$\pm$1.5 & 75.7\tiny$\pm$5.2 & 81.8\tiny$\pm$0.54 & 79.3\tiny$\pm$3.4 & 81.0\tiny$\pm$2.2 & 75.2\tiny$\pm$0.16 & - & 77.2\tiny$\pm$2.2 & - \\
BNM (V) & 74.6\tiny$\pm$3.2 & 80.5\tiny$\pm$0.89 & 81.7\tiny$\pm$1.2 & 80.8\tiny$\pm$0.69 & 82.0\tiny$\pm$0.93 & 81.5\tiny$\pm$1.4 & 81.8\tiny$\pm$0.52 & 80.2\tiny$\pm$1.4 & 81.6\tiny$\pm$1.5 & 79.1\tiny$\pm$1.1 & - & 81.8\tiny$\pm$0.63 & - \\
ClassAMI & 68.0\tiny$\pm$13 & 79.4\tiny$\pm$0.18 & 82.1\tiny$\pm$0.24 & 70.8\tiny$\pm$7.9 & 80.3\tiny$\pm$1.6 & 78.5\tiny$\pm$3.9 & 78.6\tiny$\pm$1.7 & 81.0\tiny$\pm$1.2 & 79.0\tiny$\pm$3.5 & 75.9\tiny$\pm$1.1 & - & 80.4\tiny$\pm$2.4 & - \\
SND & 70.6\tiny$\pm$3.4 & 72.2\tiny$\pm$6.5 & 77.0\tiny$\pm$0.95 & 72.6\tiny$\pm$2.0 & 76.9\tiny$\pm$1.5 & 68.5\tiny$\pm$4.6 & 72.9\tiny$\pm$3.4 & 73.8\tiny$\pm$4.5 & 73.9\tiny$\pm$3.8 & 75.2\tiny$\pm$2.5 & - & 76.1\tiny$\pm$0.68 & - \\
MixVal & 73.2\tiny$\pm$3.6 & 80.5\tiny$\pm$0.01 & 81.4\tiny$\pm$0.81 & 75.2\tiny$\pm$2.9 & 80.7\tiny$\pm$1.8 & 75.8\tiny$\pm$2.4 & 71.2\tiny$\pm$14 & 81.0\tiny$\pm$1.8 & 80.2\tiny$\pm$2.7 & 76.4\tiny$\pm$1.0 & - & 66.6\tiny$\pm$15 & - \\
TransScore & 66.7\tiny$\pm$0.80 & 79.8\tiny$\pm$0.69 & 82.2\tiny$\pm$1.1 & 79.3\tiny$\pm$1.2 & 81.5\tiny$\pm$1.5 & 81.5\tiny$\pm$1.4 & 81.3\tiny$\pm$0.38 & 70.6\tiny$\pm$12 & 81.8\tiny$\pm$0.16 & 77.1\tiny$\pm$1.6 & - & 81.4\tiny$\pm$0.79 & - \\
\hline
\end{tabular}%
}
\end{table*}

\begin{table*}[h]
\centering
\caption{Complete UDA results on CRD$\rightarrow$LDD (target accuracy, \%; mean with std). Each cell reports the mean target accuracy (std) of the checkpoint a validator selects for an algorithm. The per-algorithm columns report selection within a single algorithm, \textbf{Avg.} is their mean, and \textbf{Across-Algo} pools the checkpoints of all algorithms and selects across them. \textit{Oracle} selects using target labels and is the actual best model, \textit{SourceOnly} is trained on source data only, and \textit{TargetOnly} is trained on labeled target data.}
\label{tab:crd-ldd-resnet}
\footnotesize
\setlength{\tabcolsep}{3pt}
\renewcommand{\arraystretch}{1.0}
\resizebox{\textwidth}{!}{%
\begin{tabular}{l | c | *{9}{c} : c | c | c}
\hline
 & \textit{SourceOnly} & {MMD} & {DANN} & {CDAN} & {DALN} & {MCC} & {BNM} & {ATDOC} & {MCD} & {CoUDA} & \textbf{Avg.} & \textbf{Across-Algo} & \textit{TargetOnly} \\
\hline
\textit{Oracle} & 76.0\tiny$\pm$2.1 & 79.5\tiny$\pm$1.6 & 76.0\tiny$\pm$7.3 & 81.0\tiny$\pm$1.5 & 78.1\tiny$\pm$2.3 & 75.7\tiny$\pm$0.52 & 83.4\tiny$\pm$1.5 & 77.2\tiny$\pm$3.9 & 78.8\tiny$\pm$3.7 & 63.6\tiny$\pm$0.60 & 77.0\tiny$\pm$5.3 & 82.9\tiny$\pm$1.2 & 98.7\tiny$\pm$0.17 \\
\hline
Source-Risk & 68.6\tiny$\pm$2.1 & 45.7\tiny$\pm$1.1 & 57.1\tiny$\pm$6.5 & 66.2\tiny$\pm$14 & 58.9\tiny$\pm$15 & 61.0\tiny$\pm$9.4 & 58.3\tiny$\pm$11 & 52.6\tiny$\pm$5.0 & 57.9\tiny$\pm$10 & 56.0\tiny$\pm$1.3 & - & 54.7\tiny$\pm$0.92 & - \\
IWCV & 69.2\tiny$\pm$7.4 & 45.2\tiny$\pm$2.5 & 51.1\tiny$\pm$2.8 & 70.0\tiny$\pm$10 & 53.0\tiny$\pm$10 & 64.1\tiny$\pm$2.5 & 58.0\tiny$\pm$15 & 42.6\tiny$\pm$5.0 & 57.5\tiny$\pm$11 & 55.4\tiny$\pm$1.4 & - & 58.4\tiny$\pm$11 & - \\
DEV & 66.2\tiny$\pm$17 & 43.9\tiny$\pm$6.8 & 60.9\tiny$\pm$14 & 63.0\tiny$\pm$7.3 & 56.4\tiny$\pm$3.1 & 55.3\tiny$\pm$10 & 51.4\tiny$\pm$3.7 & 37.8\tiny$\pm$11 & 64.2\tiny$\pm$1.9 & 53.0\tiny$\pm$0.30 & - & 56.1\tiny$\pm$5.9 & - \\
DEV-N & 68.6\tiny$\pm$2.1 & 46.7\tiny$\pm$1.5 & 61.9\tiny$\pm$15 & 63.5\tiny$\pm$15 & 58.9\tiny$\pm$15 & 61.0\tiny$\pm$9.4 & 56.6\tiny$\pm$13 & 52.6\tiny$\pm$5.0 & 57.9\tiny$\pm$10 & 56.0\tiny$\pm$1.3 & - & 54.7\tiny$\pm$0.92 & - \\
Entropy & 68.5\tiny$\pm$2.7 & 45.8\tiny$\pm$1.4 & 49.2\tiny$\pm$1.7 & 73.9\tiny$\pm$13 & 60.7\tiny$\pm$21 & 65.0\tiny$\pm$2.5 & 50.1\tiny$\pm$0.09 & 42.5\tiny$\pm$5.4 & 63.2\tiny$\pm$11 & 55.4\tiny$\pm$1.8 & - & 56.5\tiny$\pm$7.5 & - \\
InfoMax & 67.5\tiny$\pm$3.4 & 45.9\tiny$\pm$1.2 & 69.7\tiny$\pm$12 & 69.9\tiny$\pm$10 & 64.5\tiny$\pm$15 & 53.4\tiny$\pm$0.49 & 49.7\tiny$\pm$1.0 & 39.5\tiny$\pm$9.6 & 63.0\tiny$\pm$12 & 55.4\tiny$\pm$1.8 & - & 57.4\tiny$\pm$11 & - \\
Corr-C & 63.4\tiny$\pm$1.7 & 58.3\tiny$\pm$12 & 70.3\tiny$\pm$12 & 63.5\tiny$\pm$10 & 64.5\tiny$\pm$15 & 55.1\tiny$\pm$0.54 & 53.8\tiny$\pm$1.7 & 49.0\tiny$\pm$12 & 56.8\tiny$\pm$11 & 55.5\tiny$\pm$2.0 & - & 66.1\tiny$\pm$13 & - \\
MCC (V) & 68.5\tiny$\pm$2.6 & 45.8\tiny$\pm$1.4 & 49.2\tiny$\pm$0.93 & 70.0\tiny$\pm$12 & 61.4\tiny$\pm$19 & 64.4\tiny$\pm$2.5 & 57.2\tiny$\pm$14 & 42.5\tiny$\pm$5.4 & 50.9\tiny$\pm$4.4 & 55.6\tiny$\pm$1.1 & - & 68.1\tiny$\pm$8.7 & - \\
BNM (V) & 68.0\tiny$\pm$2.4 & 45.9\tiny$\pm$1.2 & 69.7\tiny$\pm$12 & 69.3\tiny$\pm$11 & 64.5\tiny$\pm$15 & 53.4\tiny$\pm$0.49 & 49.7\tiny$\pm$1.0 & 39.5\tiny$\pm$9.7 & 63.0\tiny$\pm$12 & 55.4\tiny$\pm$1.8 & - & 57.4\tiny$\pm$11 & - \\
ClassAMI & 68.6\tiny$\pm$3.2 & 44.3\tiny$\pm$1.6 & 59.3\tiny$\pm$18 & 66.7\tiny$\pm$8.0 & 59.5\tiny$\pm$19 & 51.6\tiny$\pm$11 & 54.8\tiny$\pm$1.8 & 43.1\tiny$\pm$6.1 & 63.9\tiny$\pm$2.4 & 51.6\tiny$\pm$1.6 & - & 74.9\tiny$\pm$8.5 & - \\
SND & 60.1\tiny$\pm$3.7 & 59.6\tiny$\pm$9.6 & 69.0\tiny$\pm$4.3 & 67.8\tiny$\pm$8.8 & 56.9\tiny$\pm$6.1 & 49.6\tiny$\pm$1.9 & 52.4\tiny$\pm$0.84 & 50.1\tiny$\pm$2.9 & 67.4\tiny$\pm$10 & 55.3\tiny$\pm$0.38 & - & 52.3\tiny$\pm$8.2 & - \\
MixVal & 66.9\tiny$\pm$2.3 & 46.5\tiny$\pm$3.3 & 63.1\tiny$\pm$14 & 66.8\tiny$\pm$15 & 61.5\tiny$\pm$16 & 50.9\tiny$\pm$4.2 & 58.0\tiny$\pm$12 & 39.0\tiny$\pm$1.2 & 67.7\tiny$\pm$3.5 & 55.2\tiny$\pm$1.7 & - & 65.8\tiny$\pm$4.8 & - \\
TransScore & 68.4\tiny$\pm$2.6 & 63.9\tiny$\pm$15 & 60.0\tiny$\pm$14 & 70.4\tiny$\pm$13 & 54.8\tiny$\pm$9.5 & 49.0\tiny$\pm$1.9 & 52.2\tiny$\pm$4.5 & 32.2\tiny$\pm$0.56 & 52.8\tiny$\pm$7.3 & 55.3\tiny$\pm$1.9 & - & 55.0\tiny$\pm$7.4 & - \\
\hline
\end{tabular}%
}
\end{table*}

\begin{table*}[h]
\centering
\caption{Complete UDA results on OCT$\rightarrow$SLO (target accuracy, \%; mean with std). Each cell reports the mean target accuracy (std) of the checkpoint a validator selects for an algorithm. The per-algorithm columns report selection within a single algorithm, \textbf{Avg.} is their mean, and \textbf{Across-Algo} pools the checkpoints of all algorithms and selects across them. \textit{Oracle} selects using target labels and is the actual best model, \textit{SourceOnly} is trained on source data only, and \textit{TargetOnly} is trained on labeled target data.}
\label{tab:oct-slo}
\footnotesize
\setlength{\tabcolsep}{3pt}
\renewcommand{\arraystretch}{1.0}
\resizebox{\textwidth}{!}{%
\begin{tabular}{l | c | *{8}{c} : c | c | c}
\hline
 & \textit{SourceOnly} & {MMD} & {DANN} & {CDAN} & {DALN} & {MCC} & {BNM} & {ATDOC} & {MCD} & \textbf{Avg.} & \textbf{Across-Algo} & \textit{TargetOnly} \\
\hline
\textit{Oracle} & 61.9\tiny$\pm$1.3 & 66.7\tiny$\pm$0.52 & 67.2\tiny$\pm$0.37 & 67.2\tiny$\pm$0.53 & 66.0\tiny$\pm$1.0 & 64.2\tiny$\pm$1.4 & 65.4\tiny$\pm$0.19 & 59.4\tiny$\pm$2.7 & 62.9\tiny$\pm$1.2 & 64.9\tiny$\pm$2.5 & 67.4\tiny$\pm$0.15 & 75.0\tiny$\pm$0.58 \\
\hline
Source-Risk & 59.5\tiny$\pm$1.1 & 63.3\tiny$\pm$1.6 & 66.1\tiny$\pm$0.67 & 64.0\tiny$\pm$0.65 & 64.3\tiny$\pm$0.92 & 57.6\tiny$\pm$3.9 & 54.2\tiny$\pm$6.0 & 39.9\tiny$\pm$3.8 & 56.2\tiny$\pm$1.1 & - & 57.6\tiny$\pm$13 & - \\
IWCV & 56.2\tiny$\pm$5.5 & 65.8\tiny$\pm$0.42 & 61.6\tiny$\pm$1.3 & 59.7\tiny$\pm$1.4 & 53.9\tiny$\pm$2.8 & 60.2\tiny$\pm$4.1 & 54.3\tiny$\pm$14 & 44.5\tiny$\pm$5.2 & 56.0\tiny$\pm$0.82 & - & 50.3\tiny$\pm$10 & - \\
DEV & 57.1\tiny$\pm$4.1 & 64.2\tiny$\pm$2.4 & 60.8\tiny$\pm$4.9 & 62.2\tiny$\pm$1.8 & 53.4\tiny$\pm$0.65 & 56.1\tiny$\pm$6.2 & 57.2\tiny$\pm$6.0 & 51.3\tiny$\pm$5.3 & 57.5\tiny$\pm$2.8 & - & 62.7\tiny$\pm$2.8 & - \\
DEV-N & 58.7\tiny$\pm$0.49 & 63.3\tiny$\pm$1.6 & 65.2\tiny$\pm$1.9 & 63.8\tiny$\pm$0.16 & 64.4\tiny$\pm$1.1 & 59.2\tiny$\pm$4.5 & 54.2\tiny$\pm$6.0 & 42.1\tiny$\pm$7.0 & 48.1\tiny$\pm$5.6 & - & 57.9\tiny$\pm$9.3 & - \\
Entropy & 58.0\tiny$\pm$3.3 & 63.2\tiny$\pm$1.5 & 66.3\tiny$\pm$0.58 & 63.6\tiny$\pm$1.6 & 64.0\tiny$\pm$0.84 & 50.4\tiny$\pm$3.9 & 50.4\tiny$\pm$0.68 & 46.8\tiny$\pm$5.5 & 49.8\tiny$\pm$6.2 & - & 50.2\tiny$\pm$3.9 & - \\
InfoMax & 59.2\tiny$\pm$1.2 & 63.7\tiny$\pm$1.9 & 65.8\tiny$\pm$0.20 & 66.4\tiny$\pm$0.58 & 64.0\tiny$\pm$0.84 & 54.8\tiny$\pm$7.9 & 55.3\tiny$\pm$8.6 & 46.3\tiny$\pm$3.8 & 49.9\tiny$\pm$5.9 & - & 49.7\tiny$\pm$4.0 & - \\
Corr-C & 58.9\tiny$\pm$1.9 & 63.8\tiny$\pm$2.0 & 65.5\tiny$\pm$1.5 & 65.9\tiny$\pm$0.74 & 64.6\tiny$\pm$0.23 & 59.7\tiny$\pm$3.0 & 54.3\tiny$\pm$7.2 & 41.4\tiny$\pm$6.7 & 55.9\tiny$\pm$2.0 & - & 57.3\tiny$\pm$7.0 & - \\
MCC (V) & 59.1\tiny$\pm$1.6 & 63.3\tiny$\pm$1.4 & 66.3\tiny$\pm$0.58 & 65.0\tiny$\pm$0.97 & 64.0\tiny$\pm$0.84 & 54.0\tiny$\pm$7.4 & 57.5\tiny$\pm$11 & 46.8\tiny$\pm$5.5 & 49.8\tiny$\pm$6.2 & - & 54.0\tiny$\pm$7.4 & - \\
BNM (V) & 59.2\tiny$\pm$1.2 & 63.4\tiny$\pm$1.6 & 66.3\tiny$\pm$0.57 & 66.4\tiny$\pm$0.58 & 64.0\tiny$\pm$0.84 & 52.2\tiny$\pm$5.3 & 55.3\tiny$\pm$8.6 & 46.3\tiny$\pm$3.8 & 49.9\tiny$\pm$5.9 & - & 49.7\tiny$\pm$4.0 & - \\
ClassAMI & 55.5\tiny$\pm$6.7 & 64.8\tiny$\pm$1.1 & 65.7\tiny$\pm$0.88 & 63.7\tiny$\pm$1.0 & 64.9\tiny$\pm$0.52 & 54.0\tiny$\pm$6.8 & 61.2\tiny$\pm$3.4 & 44.0\tiny$\pm$9.7 & 53.0\tiny$\pm$10 & - & 50.2\tiny$\pm$5.9 & - \\
SND & 58.9\tiny$\pm$1.8 & 62.7\tiny$\pm$1.0 & 62.1\tiny$\pm$0.93 & 61.6\tiny$\pm$2.9 & 59.7\tiny$\pm$0.93 & 60.8\tiny$\pm$3.8 & 53.9\tiny$\pm$13 & 53.5\tiny$\pm$11 & 57.0\tiny$\pm$7.9 & - & 59.4\tiny$\pm$0.87 & - \\
MixVal & 59.0\tiny$\pm$1.6 & 63.3\tiny$\pm$1.4 & 65.4\tiny$\pm$0.68 & 65.3\tiny$\pm$1.0 & 64.4\tiny$\pm$0.67 & 58.2\tiny$\pm$4.2 & 52.5\tiny$\pm$3.0 & 45.4\tiny$\pm$5.5 & 53.4\tiny$\pm$2.8 & - & 51.5\tiny$\pm$3.7 & - \\
TransScore & 59.2\tiny$\pm$1.2 & 64.4\tiny$\pm$0.54 & 65.6\tiny$\pm$1.6 & 65.2\tiny$\pm$0.94 & 64.3\tiny$\pm$0.08 & 56.8\tiny$\pm$4.2 & 46.0\tiny$\pm$0.13 & 47.2\tiny$\pm$5.1 & 52.9\tiny$\pm$3.7 & - & 54.5\tiny$\pm$3.8 & - \\
\hline
\end{tabular}%
}
\end{table*}

\begin{table*}[h]
\centering
\caption{Complete UDA results on SLO$\rightarrow$OCT (target accuracy, \%; mean with std). Each cell reports the mean target accuracy (std) of the checkpoint a validator selects for an algorithm. The per-algorithm columns report selection within a single algorithm, \textbf{Avg.} is their mean, and \textbf{Across-Algo} pools the checkpoints of all algorithms and selects across them. \textit{Oracle} selects using target labels and is the actual best model, \textit{SourceOnly} is trained on source data only, and \textit{TargetOnly} is trained on labeled target data.}
\label{tab:slo-oct}
\footnotesize
\setlength{\tabcolsep}{3pt}
\renewcommand{\arraystretch}{1.0}
\resizebox{\textwidth}{!}{%
\begin{tabular}{l | c | *{8}{c} : c | c | c}
\hline
 & \textit{SourceOnly} & {MMD} & {DANN} & {CDAN} & {DALN} & {MCC} & {BNM} & {ATDOC} & {MCD} & \textbf{Avg.} & \textbf{Across-Algo} & \textit{TargetOnly} \\
\hline
\textit{Oracle} & 60.7\tiny$\pm$1.2 & 63.3\tiny$\pm$1.3 & 63.8\tiny$\pm$0.32 & 65.4\tiny$\pm$0.46 & 63.6\tiny$\pm$1.1 & 63.0\tiny$\pm$0.06 & 60.5\tiny$\pm$4.2 & 57.5\tiny$\pm$7.0 & 62.0\tiny$\pm$1.6 & 62.4\tiny$\pm$2.3 & 65.6\tiny$\pm$0.62 & 74.4\tiny$\pm$0.38 \\
\hline
Source-Risk & 57.9\tiny$\pm$1.4 & 61.1\tiny$\pm$0.12 & 62.0\tiny$\pm$0.53 & 62.5\tiny$\pm$0.28 & 59.6\tiny$\pm$1.3 & 55.7\tiny$\pm$3.7 & 47.4\tiny$\pm$2.0 & 45.0\tiny$\pm$0.76 & 54.6\tiny$\pm$6.7 & - & 55.2\tiny$\pm$6.7 & - \\
IWCV & 56.6\tiny$\pm$0.78 & 56.0\tiny$\pm$1.1 & 60.5\tiny$\pm$3.5 & 60.7\tiny$\pm$0.29 & 58.7\tiny$\pm$0.39 & 55.6\tiny$\pm$3.6 & 49.1\tiny$\pm$1.3 & 49.2\tiny$\pm$1.5 & 55.7\tiny$\pm$4.7 & - & 52.0\tiny$\pm$5.4 & - \\
DEV & 58.4\tiny$\pm$1.7 & 56.8\tiny$\pm$4.8 & 62.0\tiny$\pm$2.8 & 61.3\tiny$\pm$0.75 & 56.1\tiny$\pm$1.9 & 57.8\tiny$\pm$1.4 & 49.0\tiny$\pm$1.3 & 46.6\tiny$\pm$1.9 & 57.5\tiny$\pm$3.0 & - & 54.2\tiny$\pm$7.8 & - \\
DEV-N & 57.0\tiny$\pm$3.0 & 62.6\tiny$\pm$2.2 & 62.0\tiny$\pm$0.53 & 61.8\tiny$\pm$2.1 & 60.3\tiny$\pm$1.7 & 56.0\tiny$\pm$3.2 & 46.8\tiny$\pm$1.2 & 42.7\tiny$\pm$2.6 & 49.7\tiny$\pm$3.8 & - & 60.1\tiny$\pm$4.9 & - \\
Entropy & 57.1\tiny$\pm$1.3 & 51.6\tiny$\pm$4.5 & 61.4\tiny$\pm$1.1 & 61.6\tiny$\pm$2.0 & 61.4\tiny$\pm$1.1 & 43.8\tiny$\pm$0.65 & 50.7\tiny$\pm$0.94 & 47.2\tiny$\pm$0.85 & 51.1\tiny$\pm$6.0 & - & 45.9\tiny$\pm$3.6 & - \\
InfoMax & 56.9\tiny$\pm$1.6 & 51.7\tiny$\pm$4.5 & 61.4\tiny$\pm$1.1 & 60.0\tiny$\pm$0.65 & 61.6\tiny$\pm$1.1 & 42.7\tiny$\pm$2.5 & 47.4\tiny$\pm$0.14 & 47.5\tiny$\pm$0.47 & 52.1\tiny$\pm$5.8 & - & 45.3\tiny$\pm$4.9 & - \\
Corr-C & 57.9\tiny$\pm$1.9 & 51.8\tiny$\pm$4.7 & 60.0\tiny$\pm$0.81 & 61.5\tiny$\pm$1.2 & 61.7\tiny$\pm$1.4 & 57.3\tiny$\pm$5.3 & 48.2\tiny$\pm$0.31 & 48.2\tiny$\pm$5.7 & 51.8\tiny$\pm$5.3 & - & 49.3\tiny$\pm$1.9 & - \\
MCC (V) & 56.9\tiny$\pm$1.5 & 51.7\tiny$\pm$4.6 & 61.4\tiny$\pm$1.1 & 60.3\tiny$\pm$0.05 & 61.4\tiny$\pm$1.1 & 43.8\tiny$\pm$0.65 & 49.7\tiny$\pm$2.3 & 47.2\tiny$\pm$0.85 & 50.8\tiny$\pm$5.6 & - & 45.5\tiny$\pm$3.1 & - \\
BNM (V) & 56.9\tiny$\pm$1.5 & 51.7\tiny$\pm$4.5 & 61.4\tiny$\pm$1.1 & 60.0\tiny$\pm$0.65 & 61.6\tiny$\pm$1.1 & 42.7\tiny$\pm$2.5 & 47.4\tiny$\pm$0.19 & 47.5\tiny$\pm$0.47 & 52.2\tiny$\pm$5.9 & - & 45.3\tiny$\pm$4.9 & - \\
ClassAMI & 57.8\tiny$\pm$2.1 & 59.5\tiny$\pm$2.0 & 60.1\tiny$\pm$3.1 & 60.7\tiny$\pm$1.2 & 61.8\tiny$\pm$1.6 & 56.2\tiny$\pm$5.7 & 47.7\tiny$\pm$1.2 & 45.1\tiny$\pm$3.7 & 52.7\tiny$\pm$6.0 & - & 46.0\tiny$\pm$5.9 & - \\
SND & 59.0\tiny$\pm$2.4 & 58.4\tiny$\pm$4.5 & 59.4\tiny$\pm$0.92 & 59.1\tiny$\pm$2.1 & 57.0\tiny$\pm$1.4 & 60.0\tiny$\pm$2.0 & 53.1\tiny$\pm$9.1 & 53.6\tiny$\pm$8.1 & 58.7\tiny$\pm$2.1 & - & 57.0\tiny$\pm$1.4 & - \\
MixVal & 57.9\tiny$\pm$2.1 & 52.4\tiny$\pm$5.6 & 61.0\tiny$\pm$1.3 & 61.0\tiny$\pm$0.99 & 59.9\tiny$\pm$1.4 & 46.3\tiny$\pm$1.7 & 47.3\tiny$\pm$1.5 & 51.2\tiny$\pm$1.2 & 53.6\tiny$\pm$5.6 & - & 52.2\tiny$\pm$6.5 & - \\
TransScore & 57.6\tiny$\pm$1.7 & 59.4\tiny$\pm$4.3 & 61.4\tiny$\pm$1.1 & 64.7\tiny$\pm$0.27 & 60.3\tiny$\pm$0.93 & 45.3\tiny$\pm$2.8 & 49.8\tiny$\pm$3.0 & 43.5\tiny$\pm$1.7 & 57.3\tiny$\pm$3.2 & - & 57.3\tiny$\pm$3.2 & - \\
\hline
\end{tabular}%
}
\end{table*}

\subsection{Full Validator Reliability Analyses for Within-Algorithm Selection}
\label{sec:validator_reliability}
This section provides the full per-scenario validator reliability analyses. Each table reports, for one cross-domain scenario, the
within-algorithm Spearman correlation ($\rho$) between each validation score and target accuracy, computed over the checkpoints of each algorithm. Positive values always indicate ranking in the intended direction. The $\rho$ values here are means across folds or random seeds, whereas the example scatter plots of the main paper show a single fold or random seed for readability, so individual cells may not match exactly. Tables~\ref{tab:spearman-brain-1-2}--\ref{tab:spearman-slo-oct} cover the eleven clinically relevant transfer scenarios. Brain MRI: ADNI-1$\rightarrow$ADNI-2 (Table~\ref{tab:spearman-brain-1-2}), ADNI-1$\rightarrow$ADNI-3 (Table~\ref{tab:spearman-brain-1-3}), ADNI-2$\rightarrow$ADNI-1 (Table~\ref{tab:spearman-brain-2-1}), ADNI-2$\rightarrow$ADNI-3 (Table~\ref{tab:spearman-brain-2-3}), ADNI-1+2$\rightarrow$AIBL (Table~\ref{tab:spearman-adni-aibl}). Chest X-ray: RSNA$\rightarrow$Child CXR (Table~\ref{tab:spearman-rsna-pedia-resnet-s}), Child CXR$\rightarrow$RSNA (Table~\ref{tab:spearman-pedia-rsna-resnet}), LDD$\rightarrow$CRD (Table~\ref{tab:spearman-ldd-crd-resnet}), CRD$\rightarrow$LDD (Table~\ref{tab:spearman-crd-ldd-resnet}). Retinal: OCT$\rightarrow$SLO (Table~\ref{tab:spearman-oct-slo}), SLO$\rightarrow$OCT (Table~\ref{tab:spearman-slo-oct}).

\begin{table*}[h]
\centering
\caption{Within-algorithm Spearman's $\rho$ on ADNI-1$\rightarrow$ADNI-2. Each cell is the mean $\rho$ over five folds between the validation score and target accuracy across checkpoints within a single algorithm. Positive values indicate ranking in the intended direction.}
\label{tab:spearman-brain-1-2}
\footnotesize
\setlength{\tabcolsep}{4pt}
\renewcommand{\arraystretch}{1.0}
\begin{tabular}{l | c c c c c c c c c}
\hline
 & {MMD} & {DANN} & {CDAN} & {DALN} & {MCC} & {BNM} & {ATDOC} & {MCD} & {AD2A} \\
\hline
Source-Risk & 0.83 & 0.76 & 0.74 & 0.79 & 0.74 & 0.73 & 0.54 & 0.79 & 0.80 \\
IWCV & 0.53 & 0.29 & 0.31 & -0.19 & 0.30 & 0.10 & 0.13 & 0.48 & 0.38 \\
DEV & 0.58 & 0.49 & 0.47 & 0.06 & 0.21 & 0.20 & 0.16 & 0.57 & 0.57 \\
DEV-N & 0.79 & 0.74 & 0.72 & 0.73 & 0.73 & 0.70 & 0.56 & 0.78 & 0.79 \\
Entropy & 0.44 & 0.61 & 0.60 & 0.62 & 0.54 & 0.54 & 0.33 & 0.72 & 0.67 \\
InfoMax & 0.80 & 0.87 & 0.87 & 0.90 & 0.80 & 0.82 & 0.63 & 0.81 & 0.89 \\
Corr-C & 0.84 & 0.79 & 0.81 & 0.80 & 0.76 & 0.76 & 0.64 & 0.82 & 0.87 \\
MCC (V) & 0.73 & 0.83 & 0.82 & 0.84 & 0.73 & 0.77 & 0.56 & 0.78 & 0.86 \\
BNM (V) & 0.76 & 0.86 & 0.86 & 0.88 & 0.78 & 0.82 & 0.62 & 0.80 & 0.88 \\
SND & 0.78 & 0.58 & 0.57 & 0.57 & 0.52 & 0.47 & 0.40 & 0.78 & 0.66 \\
ClassAMI & 0.29 & 0.70 & 0.70 & 0.72 & 0.71 & 0.70 & 0.61 & 0.47 & 0.71 \\
MixVal & 0.31 & 0.38 & 0.38 & 0.25 & 0.33 & 0.33 & 0.29 & 0.31 & 0.41 \\
TransScore & 0.75 & 0.86 & 0.85 & 0.86 & 0.79 & 0.79 & 0.61 & 0.78 & 0.88 \\
\hline
\end{tabular}%
\end{table*}

\begin{table*}[h]
\centering
\caption{Within-algorithm Spearman's $\rho$ on ADNI-1$\rightarrow$ADNI-3. Each cell is the mean $\rho$ over five folds between the validation score and target accuracy across checkpoints within a single algorithm. Positive values indicate ranking in the intended direction.}
\label{tab:spearman-brain-1-3}
\footnotesize
\setlength{\tabcolsep}{4pt}
\renewcommand{\arraystretch}{1.0}
\begin{tabular}{l | c c c c c c c c c}
\hline
 & {MMD} & {DANN} & {CDAN} & {DALN} & {MCC} & {BNM} & {ATDOC} & {MCD} & {AD2A} \\
\hline
Source-Risk & 0.52 & 0.52 & 0.61 & 0.40 & 0.50 & 0.40 & 0.47 & 0.73 & 0.75 \\
IWCV & 0.16 & 0.14 & 0.22 & -0.12 & 0.28 & 0.11 & 0.09 & 0.52 & 0.36 \\
DEV & 0.22 & 0.26 & 0.35 & 0.11 & 0.29 & 0.22 & -0.07 & 0.54 & 0.48 \\
DEV-N & 0.51 & 0.52 & 0.61 & 0.39 & 0.49 & 0.40 & 0.24 & 0.69 & 0.75 \\
Entropy & 0.35 & 0.53 & 0.44 & 0.22 & 0.32 & 0.24 & 0.05 & 0.51 & 0.56 \\
InfoMax & 0.46 & 0.61 & 0.58 & 0.39 & 0.33 & 0.18 & 0.54 & 0.65 & 0.73 \\
Corr-C & 0.40 & 0.35 & 0.37 & 0.21 & 0.15 & 0.03 & 0.39 & 0.64 & 0.64 \\
MCC (V) & 0.45 & 0.66 & 0.64 & 0.41 & 0.43 & 0.30 & 0.52 & 0.64 & 0.70 \\
BNM (V) & 0.46 & 0.63 & 0.61 & 0.40 & 0.35 & 0.20 & 0.53 & 0.63 & 0.72 \\
SND & 0.34 & 0.14 & 0.24 & 0.18 & 0.12 & -0.01 & 0.28 & 0.61 & 0.52 \\
ClassAMI & 0.02 & 0.56 & 0.58 & 0.37 & 0.50 & 0.41 & 0.45 & 0.47 & 0.61 \\
MixVal & 0.11 & 0.32 & 0.21 & 0.15 & 0.22 & 0.16 & 0.34 & 0.36 & 0.34 \\
TransScore & 0.42 & 0.60 & 0.59 & 0.37 & 0.32 & 0.16 & 0.52 & 0.54 & 0.71 \\
\hline
\end{tabular}%
\end{table*}

\begin{table*}[h]
\centering
\caption{Within-algorithm Spearman's $\rho$ on ADNI-2$\rightarrow$ADNI-1. Each cell is the mean $\rho$ over five folds between the validation score and target accuracy across checkpoints within a single algorithm. Positive values indicate ranking in the intended direction.}
\label{tab:spearman-brain-2-1}
\footnotesize
\setlength{\tabcolsep}{4pt}
\renewcommand{\arraystretch}{1.0}
\begin{tabular}{l | c c c c c c c c c}
\hline
 & {MMD} & {DANN} & {CDAN} & {DALN} & {MCC} & {BNM} & {ATDOC} & {MCD} & {AD2A} \\
\hline
Source-Risk & 0.72 & 0.51 & 0.53 & 0.59 & 0.58 & 0.57 & 0.56 & 0.74 & 0.79 \\
IWCV & 0.38 & 0.15 & 0.10 & -0.28 & 0.39 & 0.15 & 0.23 & 0.31 & 0.33 \\
DEV & 0.42 & 0.14 & 0.33 & 0.18 & 0.30 & 0.16 & 0.21 & 0.41 & 0.47 \\
DEV-N & 0.69 & 0.54 & 0.57 & 0.62 & 0.61 & 0.62 & 0.62 & 0.76 & 0.81 \\
Entropy & 0.34 & 0.45 & 0.44 & 0.44 & 0.33 & 0.35 & 0.39 & 0.67 & 0.58 \\
InfoMax & 0.75 & 0.69 & 0.69 & 0.78 & 0.70 & 0.71 & 0.67 & 0.78 & 0.89 \\
Corr-C & 0.78 & 0.78 & 0.60 & 0.71 & 0.66 & 0.65 & 0.69 & 0.82 & 0.90 \\
MCC (V) & 0.68 & 0.64 & 0.65 & 0.74 & 0.55 & 0.63 & 0.60 & 0.76 & 0.86 \\
BNM (V) & 0.73 & 0.69 & 0.69 & 0.77 & 0.70 & 0.71 & 0.66 & 0.77 & 0.87 \\
SND & 0.61 & 0.36 & 0.18 & 0.37 & 0.43 & 0.44 & 0.33 & 0.78 & 0.66 \\
ClassAMI & 0.29 & 0.65 & 0.59 & 0.66 & 0.62 & 0.62 & 0.58 & 0.48 & 0.75 \\
MixVal & 0.37 & 0.36 & 0.36 & 0.40 & 0.41 & 0.41 & 0.45 & 0.26 & 0.44 \\
TransScore & 0.73 & 0.70 & 0.67 & 0.76 & 0.69 & 0.68 & 0.61 & 0.76 & 0.87 \\
\hline
\end{tabular}%
\end{table*}

\begin{table*}[h]
\centering
\caption{Within-algorithm Spearman's $\rho$ on ADNI-2$\rightarrow$ADNI-3. Each cell is the mean $\rho$ over five folds between the validation score and target accuracy across checkpoints within a single algorithm. Positive values indicate ranking in the intended direction.}
\label{tab:spearman-brain-2-3}
\footnotesize
\setlength{\tabcolsep}{4pt}
\renewcommand{\arraystretch}{1.0}
\begin{tabular}{l | c c c c c c c c c}
\hline
 & {MMD} & {DANN} & {CDAN} & {DALN} & {MCC} & {BNM} & {ATDOC} & {MCD} & {AD2A} \\
\hline
Source-Risk & 0.72 & 0.72 & 0.66 & 0.69 & 0.67 & 0.56 & 0.53 & 0.76 & 0.82 \\
IWCV & 0.41 & 0.26 & 0.23 & -0.23 & 0.20 & 0.01 & 0.24 & 0.54 & 0.60 \\
DEV & 0.48 & 0.47 & 0.36 & 0.18 & 0.36 & 0.26 & 0.36 & 0.71 & 0.68 \\
DEV-N & 0.66 & 0.64 & 0.57 & 0.58 & 0.58 & 0.47 & 0.52 & 0.73 & 0.78 \\
Entropy & 0.27 & 0.28 & 0.15 & 0.17 & 0.19 & 0.23 & 0.19 & 0.64 & 0.49 \\
InfoMax & 0.74 & 0.70 & 0.68 & 0.68 & 0.64 & 0.50 & 0.36 & 0.64 & 0.80 \\
Corr-C & 0.62 & 0.46 & 0.55 & 0.56 & 0.48 & 0.31 & 0.18 & 0.58 & 0.58 \\
MCC (V) & 0.69 & 0.67 & 0.60 & 0.62 & 0.60 & 0.57 & 0.41 & 0.70 & 0.73 \\
BNM (V) & 0.74 & 0.72 & 0.68 & 0.69 & 0.65 & 0.54 & -0.04 & 0.53 & 0.78 \\
SND & 0.53 & 0.35 & 0.45 & 0.49 & 0.35 & 0.21 & 0.17 & 0.52 & 0.44 \\
ClassAMI & -0.01 & 0.66 & 0.57 & 0.56 & 0.65 & 0.58 & 0.55 & 0.25 & 0.55 \\
MixVal & 0.39 & 0.28 & 0.28 & 0.24 & 0.29 & 0.28 & 0.23 & 0.16 & 0.33 \\
TransScore & 0.73 & 0.71 & 0.69 & 0.67 & 0.64 & 0.52 & 0.39 & 0.65 & 0.78 \\
\hline
\end{tabular}%
\end{table*}

\begin{table*}[h]
\centering
\caption{Within-algorithm Spearman's $\rho$ on ADNI-1+2$\rightarrow$AIBL. Each cell is the mean $\rho$ over five folds between the validation score and target accuracy across checkpoints within a single algorithm. Positive values indicate ranking in the intended direction.}
\label{tab:spearman-adni-aibl}
\footnotesize
\setlength{\tabcolsep}{4pt}
\renewcommand{\arraystretch}{1.0}
\begin{tabular}{l | c c c c c c c c c}
\hline
 & {MMD} & {DANN} & {CDAN} & {DALN} & {MCC} & {BNM} & {ATDOC} & {MCD} & {AD2A} \\
\hline
Source-Risk & 0.71 & 0.56 & 0.65 & 0.57 & 0.62 & 0.54 & 0.54 & 0.65 & 0.82 \\
IWCV & 0.52 & 0.15 & 0.18 & -0.29 & 0.25 & 0.08 & 0.22 & 0.41 & 0.64 \\
DEV & 0.60 & 0.45 & 0.37 & 0.24 & 0.43 & 0.30 & 0.43 & 0.52 & 0.72 \\
DEV-N & 0.71 & 0.56 & 0.62 & 0.57 & 0.62 & 0.56 & 0.57 & 0.66 & 0.81 \\
Entropy & 0.55 & 0.46 & 0.46 & 0.41 & 0.50 & 0.41 & 0.38 & 0.71 & 0.74 \\
InfoMax & 0.65 & 0.40 & 0.58 & 0.53 & 0.45 & 0.31 & 0.53 & 0.74 & 0.79 \\
Corr-C & 0.42 & -0.04 & 0.24 & 0.13 & 0.04 & -0.12 & -0.09 & 0.48 & 0.38 \\
MCC (V) & 0.65 & 0.54 & 0.62 & 0.57 & 0.58 & 0.44 & 0.44 & 0.72 & 0.81 \\
BNM (V) & 0.66 & 0.47 & 0.61 & 0.55 & 0.50 & 0.35 & -0.17 & 0.54 & 0.81 \\
SND & 0.34 & -0.14 & 0.09 & 0.08 & -0.11 & -0.20 & -0.23 & 0.31 & 0.16 \\
ClassAMI & -0.25 & 0.37 & 0.38 & 0.36 & 0.46 & 0.50 & 0.38 & 0.31 & 0.72 \\
MixVal & 0.22 & 0.20 & 0.26 & 0.16 & 0.05 & 0.11 & 0.13 & 0.22 & 0.34 \\
TransScore & 0.60 & 0.42 & 0.57 & 0.52 & 0.42 & 0.29 & 0.29 & 0.65 & 0.79 \\
\hline
\end{tabular}%
\end{table*}

\begin{table*}[h]
\centering
\caption{Within-algorithm Spearman's $\rho$ on RSNA$\rightarrow$Child CXR. Each cell is the mean $\rho$ over five folds between the validation score and target accuracy across checkpoints within a single algorithm. Positive values indicate ranking in the intended direction.}
\label{tab:spearman-rsna-pedia-resnet-s}
\footnotesize
\setlength{\tabcolsep}{4pt}
\renewcommand{\arraystretch}{1.0}
\begin{tabular}{l | c c c c c c c c c}
\hline
 & {MMD} & {DANN} & {CDAN} & {DALN} & {MCC} & {BNM} & {ATDOC} & {MCD} & {CoUDA} \\
\hline
Source-Risk & 0.09 & 0.29 & -0.01 & 0.21 & 0.48 & 0.29 & -0.27 & 0.52 & 0.63 \\
IWCV & -0.01 & 0.13 & -0.08 & -0.02 & 0.41 & 0.18 & 0.03 & 0.41 & 0.42 \\
DEV & -0.01 & 0.07 & -0.17 & -0.13 & 0.22 & 0.06 & 0.09 & 0.04 & 0.21 \\
DEV-N & -0.24 & -0.20 & -0.25 & 0.20 & 0.37 & 0.25 & -0.18 & 0.26 & 0.47 \\
Entropy & -0.17 & -0.13 & -0.37 & -0.20 & 0.31 & 0.11 & -0.26 & 0.30 & 0.44 \\
InfoMax & 0.27 & 0.39 & 0.03 & 0.05 & 0.66 & 0.55 & -0.07 & 0.57 & 0.61 \\
Corr-C & 0.76 & 0.77 & 0.62 & 0.13 & 0.60 & 0.56 & -0.01 & 0.74 & 0.71 \\
MCC (V) & 0.08 & 0.16 & -0.19 & -0.06 & 0.44 & 0.26 & -0.31 & 0.37 & 0.55 \\
BNM (V) & 0.22 & 0.34 & -0.01 & 0.04 & 0.66 & 0.54 & -0.08 & 0.54 & 0.61 \\
SND & 0.76 & 0.76 & 0.65 & 0.18 & 0.42 & 0.48 & 0.15 & 0.64 & 0.63 \\
ClassAMI & -0.17 & 0.68 & 0.48 & -0.26 & 0.57 & 0.46 & -0.07 & 0.19 & -0.36 \\
MixVal & 0.49 & 0.52 & 0.43 & 0.41 & 0.24 & 0.06 & -0.15 & 0.39 & 0.04 \\
TransScore & 0.05 & 0.13 & -0.09 & 0.15 & 0.63 & 0.57 & -0.10 & 0.24 & 0.62 \\
\hline
\end{tabular}%
\end{table*}

\begin{table*}[h]
\centering
\caption{Within-algorithm Spearman's $\rho$ on Child CXR$\rightarrow$RSNA. Each cell is the mean $\rho$ over five folds between the validation score and target accuracy across checkpoints within a single algorithm. Positive values indicate ranking in the intended direction.}
\label{tab:spearman-pedia-rsna-resnet}
\footnotesize
\setlength{\tabcolsep}{4pt}
\renewcommand{\arraystretch}{1.0}
\begin{tabular}{l | c c c c c c c c c}
\hline
 & {MMD} & {DANN} & {CDAN} & {DALN} & {MCC} & {BNM} & {ATDOC} & {MCD} & {CoUDA} \\
\hline
Source-Risk & 0.20 & 0.15 & -0.09 & 0.22 & 0.15 & 0.17 & -0.06 & 0.01 & -0.12 \\
IWCV & 0.24 & 0.18 & 0.14 & 0.14 & 0.18 & 0.09 & -0.11 & 0.18 & 0.19 \\
DEV & 0.21 & 0.20 & 0.13 & 0.15 & 0.19 & 0.10 & -0.13 & 0.11 & 0.20 \\
DEV-N & 0.21 & 0.14 & -0.09 & 0.22 & 0.15 & 0.18 & -0.07 & 0.02 & -0.11 \\
Entropy & -0.09 & -0.29 & -0.26 & 0.44 & 0.08 & 0.32 & -0.18 & 0.69 & -0.46 \\
InfoMax & 0.81 & 0.78 & 0.88 & 0.47 & 0.55 & 0.44 & -0.04 & 0.75 & -0.37 \\
Corr-C & 0.95 & 0.98 & 0.99 & 0.43 & 0.58 & 0.38 & 0.03 & 0.67 & -0.17 \\
MCC (V) & 0.01 & -0.15 & 0.05 & 0.42 & 0.15 & 0.34 & -0.18 & 0.68 & -0.46 \\
BNM (V) & 0.77 & 0.71 & 0.85 & 0.47 & 0.53 & 0.44 & -0.05 & 0.75 & -0.38 \\
SND & 0.91 & 0.93 & 0.97 & -0.27 & 0.56 & 0.29 & 0.16 & 0.39 & 0.57 \\
ClassAMI & 0.28 & -0.10 & 0.18 & 0.21 & 0.04 & 0.12 & 0.09 & 0.47 & 0.02 \\
MixVal & 0.11 & 0.25 & 0.45 & 0.40 & -0.13 & -0.14 & 0.08 & 0.16 & -0.23 \\
TransScore & 0.21 & -0.04 & -0.11 & 0.18 & 0.37 & 0.39 & -0.17 & 0.71 & -0.41 \\
\hline
\end{tabular}%
\end{table*}

\begin{table*}[h]
\centering
\caption{Within-algorithm Spearman's $\rho$ on LDD$\rightarrow$CRD. Each cell is the mean $\rho$ over five folds between the validation score and target accuracy across checkpoints within a single algorithm. Positive values indicate ranking in the intended direction.}
\label{tab:spearman-ldd-crd-resnet}
\footnotesize
\setlength{\tabcolsep}{4pt}
\renewcommand{\arraystretch}{1.0}
\begin{tabular}{l | c c c c c c c c c}
\hline
 & {MMD} & {DANN} & {CDAN} & {DALN} & {MCC} & {BNM} & {ATDOC} & {MCD} & {CoUDA} \\
\hline
Source-Risk & 0.56 & 0.54 & 0.04 & 0.69 & 0.20 & 0.74 & 0.27 & 0.63 & -0.03 \\
IWCV & 0.41 & 0.16 & -0.23 & -0.07 & -0.18 & -0.33 & -0.01 & -0.15 & -0.17 \\
DEV & 0.43 & 0.18 & -0.13 & 0.13 & -0.09 & -0.12 & 0.06 & 0.17 & -0.09 \\
DEV-N & 0.55 & 0.49 & 0.02 & 0.67 & 0.18 & 0.73 & 0.27 & 0.57 & -0.06 \\
Entropy & 0.55 & 0.56 & -0.08 & 0.67 & 0.07 & 0.65 & 0.27 & 0.69 & -0.06 \\
InfoMax & 0.76 & 0.84 & 0.86 & 0.87 & 0.89 & 0.84 & 0.41 & 0.93 & 0.44 \\
Corr-C & 0.67 & 0.76 & 0.88 & 0.85 & 0.87 & 0.78 & 0.43 & 0.85 & 0.82 \\
MCC (V) & 0.62 & 0.63 & 0.07 & 0.77 & 0.20 & 0.77 & 0.35 & 0.78 & 0.06 \\
BNM (V) & 0.74 & 0.82 & 0.82 & 0.86 & 0.87 & 0.84 & 0.41 & 0.93 & 0.39 \\
SND & 0.14 & 0.32 & 0.72 & 0.13 & 0.74 & 0.37 & 0.12 & 0.51 & 0.70 \\
ClassAMI & 0.30 & 0.11 & 0.24 & 0.41 & 0.26 & 0.68 & 0.46 & 0.50 & 0.43 \\
MixVal & 0.60 & 0.56 & 0.39 & 0.67 & 0.21 & 0.02 & 0.27 & 0.45 & 0.39 \\
TransScore & 0.44 & 0.56 & 0.21 & 0.67 & 0.73 & 0.80 & 0.26 & 0.77 & 0.08 \\
\hline
\end{tabular}%
\end{table*}

\begin{table*}[h]
\centering
\caption{Within-algorithm Spearman's $\rho$ on CRD$\rightarrow$LDD. Each cell is the mean $\rho$ over five folds between the validation score and target accuracy across checkpoints within a single algorithm. Positive values indicate ranking in the intended direction.}
\label{tab:spearman-crd-ldd-resnet}
\footnotesize
\setlength{\tabcolsep}{4pt}
\renewcommand{\arraystretch}{1.0}
\begin{tabular}{l | c c c c c c c c c}
\hline
 & {MMD} & {DANN} & {CDAN} & {DALN} & {MCC} & {BNM} & {ATDOC} & {MCD} & {CoUDA} \\
\hline
Source-Risk & 0.09 & 0.35 & 0.42 & 0.22 & 0.48 & 0.22 & -0.08 & -0.15 & 0.24 \\
IWCV & 0.31 & 0.04 & 0.19 & 0.22 & 0.23 & 0.05 & 0.07 & 0.10 & 0.17 \\
DEV & 0.24 & 0.14 & 0.24 & 0.27 & 0.14 & 0.05 & 0.02 & 0.14 & 0.16 \\
DEV-N & 0.11 & 0.36 & 0.43 & 0.23 & 0.48 & 0.22 & -0.07 & -0.14 & 0.26 \\
Entropy & -0.13 & 0.23 & 0.28 & 0.13 & 0.49 & 0.04 & -0.16 & -0.18 & 0.19 \\
InfoMax & 0.08 & 0.38 & 0.47 & 0.29 & -0.17 & 0.34 & -0.02 & -0.30 & 0.17 \\
Corr-C & 0.45 & 0.51 & 0.43 & 0.49 & -0.28 & 0.39 & 0.34 & -0.40 & 0.06 \\
MCC (V) & -0.04 & 0.26 & 0.34 & 0.20 & 0.49 & 0.10 & -0.15 & -0.19 & 0.19 \\
BNM (V) & 0.06 & 0.37 & 0.45 & 0.28 & -0.15 & 0.32 & -0.02 & -0.29 & 0.17 \\
SND & 0.53 & 0.20 & 0.15 & 0.30 & -0.35 & 0.35 & 0.39 & -0.18 & -0.17 \\
ClassAMI & -0.06 & 0.33 & 0.34 & 0.24 & 0.39 & 0.16 & -0.05 & -0.16 & -0.16 \\
MixVal & -0.08 & 0.17 & 0.28 & 0.10 & 0.28 & 0.09 & -0.17 & 0.01 & 0.05 \\
TransScore & -0.07 & 0.19 & 0.28 & 0.18 & -0.10 & 0.23 & -0.09 & -0.22 & 0.17 \\
\hline
\end{tabular}%
\end{table*}

\begin{table*}[h]
\centering
\caption{Within-algorithm Spearman's $\rho$ on OCT$\rightarrow$SLO. Each cell is the mean $\rho$ over three random seeds between the validation score and target accuracy across checkpoints within a single algorithm. Positive values indicate ranking in the intended direction.}
\label{tab:spearman-oct-slo}
\footnotesize
\setlength{\tabcolsep}{4pt}
\renewcommand{\arraystretch}{1.0}
\begin{tabular}{l | c c c c c c c c}
\hline
 & {MMD} & {DANN} & {CDAN} & {DALN} & {MCC} & {BNM} & {ATDOC} & {MCD} \\
\hline
Source-Risk & -0.08 & 0.31 & 0.27 & 0.63 & -0.15 & -0.20 & -0.29 & 0.13 \\
IWCV & 0.06 & -0.15 & -0.06 & -0.64 & 0.22 & -0.03 & 0.07 & -0.14 \\
DEV & 0.02 & -0.07 & 0.02 & -0.47 & 0.16 & -0.13 & 0.17 & -0.21 \\
DEV-N & -0.05 & 0.37 & 0.21 & 0.62 & -0.16 & -0.19 & -0.27 & 0.00 \\
Entropy & -0.15 & 0.34 & 0.22 & 0.75 & -0.11 & -0.19 & -0.32 & 0.21 \\
InfoMax & -0.11 & 0.39 & 0.60 & 0.77 & 0.45 & -0.09 & -0.38 & 0.31 \\
Corr-C & 0.15 & 0.46 & 0.70 & 0.79 & 0.52 & 0.00 & -0.30 & 0.35 \\
MCC (V) & -0.14 & 0.35 & 0.26 & 0.77 & 0.00 & -0.18 & -0.33 & 0.24 \\
BNM (V) & -0.11 & 0.39 & 0.56 & 0.78 & 0.44 & -0.09 & -0.37 & 0.31 \\
SND & 0.26 & -0.04 & 0.14 & 0.19 & 0.51 & 0.18 & -0.09 & 0.29 \\
ClassAMI & 0.12 & 0.35 & 0.29 & 0.24 & 0.18 & 0.12 & -0.06 & -0.10 \\
MixVal & -0.12 & 0.35 & 0.30 & 0.70 & 0.25 & -0.04 & -0.31 & 0.19 \\
TransScore & -0.16 & 0.18 & 0.16 & 0.69 & 0.19 & -0.26 & -0.32 & 0.28 \\
\hline
\end{tabular}%
\end{table*}

\begin{table*}[h]
\centering
\caption{Within-algorithm Spearman's $\rho$ on SLO$\rightarrow$OCT. Each cell is the mean $\rho$ over three random seeds between the validation score and target accuracy across checkpoints within a single algorithm. Positive values indicate ranking in the intended direction.}
\label{tab:spearman-slo-oct}
\footnotesize
\setlength{\tabcolsep}{4pt}
\renewcommand{\arraystretch}{1.0}
\begin{tabular}{l | c c c c c c c c}
\hline
 & {MMD} & {DANN} & {CDAN} & {DALN} & {MCC} & {BNM} & {ATDOC} & {MCD} \\
\hline
Source-Risk & 0.22 & 0.26 & 0.50 & 0.53 & -0.33 & -0.11 & -0.30 & 0.19 \\
IWCV & 0.02 & -0.06 & -0.13 & -0.40 & 0.17 & 0.12 & 0.21 & 0.04 \\
DEV & -0.15 & 0.17 & 0.07 & 0.00 & 0.04 & 0.08 & 0.04 & -0.04 \\
DEV-N & 0.16 & 0.24 & 0.39 & 0.54 & -0.30 & -0.08 & -0.30 & 0.18 \\
Entropy & 0.07 & 0.24 & 0.53 & 0.65 & -0.39 & -0.14 & -0.32 & 0.13 \\
InfoMax & 0.11 & 0.25 & 0.58 & 0.68 & -0.12 & -0.05 & -0.38 & 0.18 \\
Corr-C & 0.04 & 0.20 & 0.38 & 0.69 & 0.02 & -0.07 & -0.30 & 0.26 \\
MCC (V) & 0.10 & 0.25 & 0.54 & 0.67 & -0.39 & -0.13 & -0.33 & 0.14 \\
BNM (V) & 0.11 & 0.25 & 0.59 & 0.68 & -0.14 & -0.06 & -0.38 & 0.18 \\
SND & 0.00 & -0.04 & -0.20 & 0.12 & 0.19 & 0.01 & -0.05 & 0.18 \\
ClassAMI & -0.39 & 0.31 & 0.46 & 0.60 & -0.18 & -0.12 & -0.19 & -0.03 \\
MixVal & 0.13 & 0.27 & 0.51 & 0.59 & -0.13 & -0.09 & -0.15 & 0.32 \\
TransScore & 0.10 & 0.19 & 0.51 & 0.17 & -0.31 & -0.05 & -0.31 & 0.24 \\
\hline
\end{tabular}%
\end{table*}




\end{document}